%% file: main.tex
\documentclass[10pt,twocolumn,letterpaper]{article}

\usepackage{multirow}
\usepackage{algorithm}
\usepackage{algorithmic}
\usepackage{amsmath}
\usepackage{subcaption}
\usepackage{multirow}
\usepackage{color}
\usepackage{bm}
\usepackage{amsfonts}
\usepackage{makecell}
\usepackage{colortbl}
\usepackage{arydshln}
\usepackage[table,xcdraw]{xcolor}
\usepackage{pifont}
\usepackage{dashbox}
\usepackage{authblk}
\usepackage{graphicx}
\usepackage{amssymb}
\usepackage{booktabs}
\usepackage{wrapfig}
\usepackage[accsupp]{axessibility}  

\usepackage{cvpr}              
\input{preamble}
\definecolor{cvprblue}{rgb}{0.21,0.49,0.74}
\usepackage[pagebackref,breaklinks,colorlinks,allcolors=cvprblue]{hyperref}

\def\paperID{***} 
\def\confName{CVPR}
\def\confYear{2026}

\title{Direct Segmentation without Logits Optimization for Training-Free Open-Vocabulary Semantic Segmentation}

\author{
\vspace{-0.5cm}
Jiahao Li$^1$, Yang Lu$^1$, Yachao Zhang$^{1\ast}$, Fangyong Wang$^2$, Yuan Xie$^3$, Yanyun Qu$^{1}$\thanks{Corresponding author. Code: https://github.com/liblacklucy/DSLO}\\
\vspace{-0.3cm}
$^1$Key Laboratory of Multimedia Trusted Perception and Efficient Computing, Ministry of Education of China, Xiamen University. $^2$Hanjiang National Laboratory. $^3$East China Normal University.\\
{\tt\small lijiahao1@stu.xmu.edu.cn} \quad {\tt\small \{luyang, yachaozhang, yyqu\}@xmu.edu.cn}
\vspace{-0.3cm}
}

\begin{document}
\maketitle
\input{sec/0_abstract}    
\input{sec/1_introduction}
\input{sec/2_relatedwork}
\input{sec/3_approach}
\input{sec/4_experiments}
\input{sec/5_conclusion}
\input{sec/6_acknowledgments}
{
    \small
    \bibliographystyle{ieeenat_fullname}
    \bibliography{main}
}

\input{sec/X_suppl}

\end{document}

%% file: sec/0_abstract.tex
\begin{abstract}
Open-vocabulary semantic segmentation (OVSS) aims to segment arbitrary category regions in images using open-vocabulary prompts, necessitating that existing methods possess pixel-level vision-language alignment capability. Typically, this capability involves computing the cosine similarity, \ie, logits, between visual and linguistic features, and minimizing the distribution discrepancy between the logits and the ground truth (GT) to generate optimal logits that are subsequently used to construct segmentation maps, yet it depends on time-consuming iterative training or model-specific attention modulation. In this work, we propose a more direct approach that eschews the logits-optimization process by directly deriving an analytic solution for the segmentation map. We posit a key hypothesis: the distribution discrepancy encodes semantic information; specifically, this discrepancy exhibits consistency across patches belonging to the same category but inconsistency across different categories. Based on this hypothesis, we directly utilize the analytic solution of this distribution discrepancy as the semantic maps. In other words, we reformulate the optimization of the distribution discrepancy as deriving its analytic solution, thereby eliminating time-consuming iterative training, freeing us from model-specific attention modulation, and achieving state-of-the-art performance on eight benchmark datasets. 

\end{abstract}

%% file: sec/1_introduction.tex
\section{Introduction}

\begin{figure}[t]
  \centering
    \includegraphics[width=\linewidth]{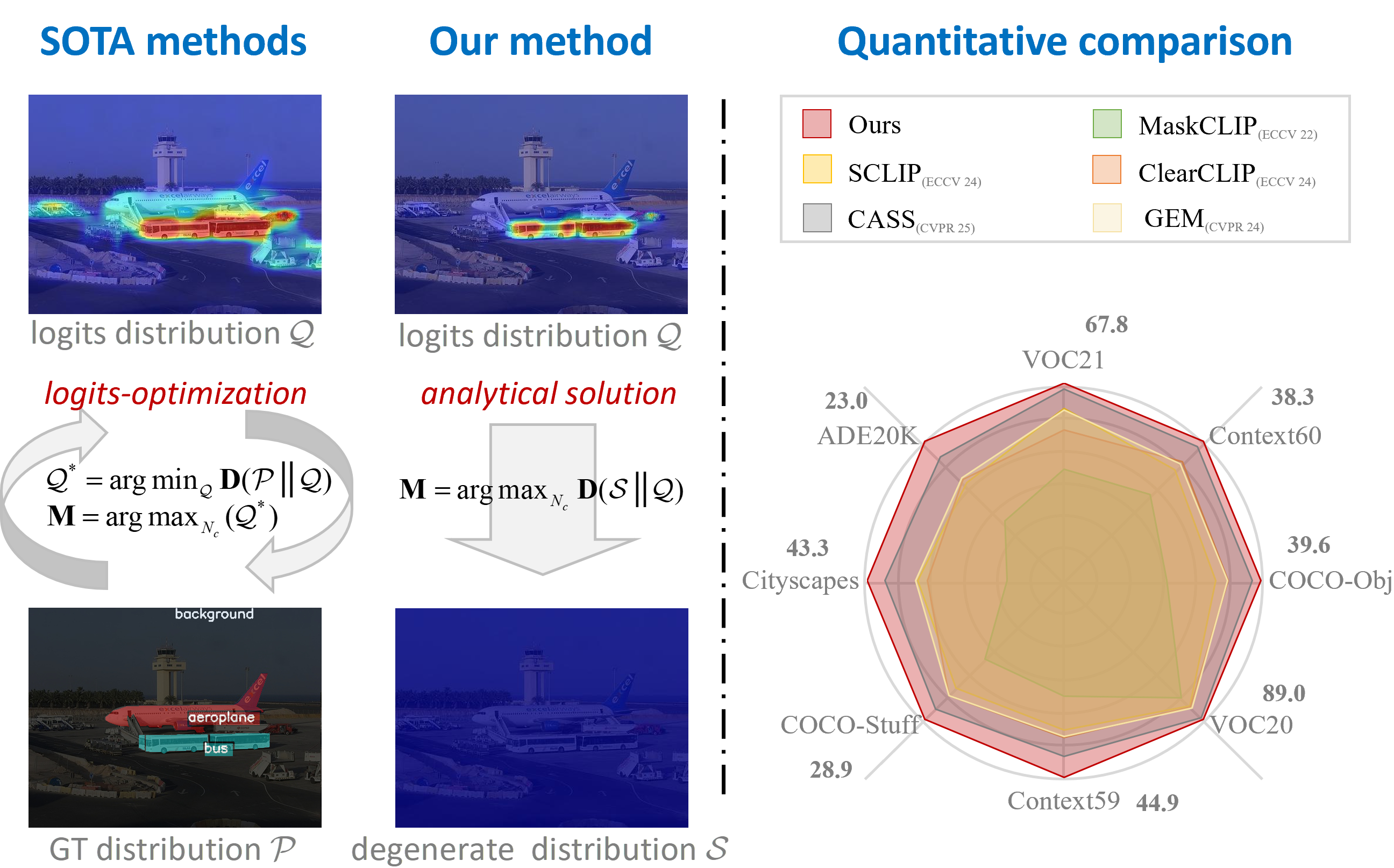}
  \caption{Compared with the logits-optimization methods seeking optimal solutions, we reformulate the problem as an analytical solution form. This reformulation confers three key advantages: independence from GT annotations, elimination of time-consuming training, and freedom from model-specific modulation.}
  \label{fig:intro}
\end{figure}

Open-vocabulary semantic segmentation (OVSS) aims to recognize the category of each pixel using class-specific textual descriptions, and has garnered significant attention~\cite{he2023primitive,zhao2017open,li2020consistent,shen2021conterfactual,li2025novel}. Typically, prevailing approaches follow an \textbf{iterative training paradigm}~\cite{zhou2022extract,liu2023delving,qin2023freeseg}: they first compute cosine similarity between visual and linguistic features (termed logits), then iteratively minimize the discrepancy between these logits and the ground truth (GT) distribution to obtain optimal logits, and finally apply an $\arg\max$ operation to the optimal logits to generate segmentation maps. Despite achieving remarkable performance, these methods rely on GT annotations and time-consuming training processes. To alleviate these limitations, the prevailing unsupervised training-free methods employ an \textbf{attention modulation paradigm}~\cite{lan2024clearclip,wang2024sclip,lan2024proxyclip}: they calibrate core self-attention computations to rectify fine-grained misalignment between visual and linguistic features, thereby constructing optimal logits for segmentation map generation. Such attention modulation techniques intrinsically perform denoising on self-attention tensor, where the noise is typically data-agnostic yet model-specific, \eg, in most CLIP-based attention modulation~\cite{wang2024sclip,hajimiri2025pay}, which results in poor generalization to other foundation models.

Both paradigms fundamentally converge on identical methodological core: prioritizing derivation of optimal logits followed by construction of segmentation maps, \ie, logits-optimization. This optimization aim to optimize the distribution discrepancy between logits and GT to obtain the optimal logits, a process that is either time-consuming or model-specific. On the contrary, we propose a more direct approach that circumvents the logits-optimization phase, eliminating the need to solve for optimal logits and directly obtaining the final segmentation maps. This design provides three key advantages: independence from GT annotations, elimination of time-consuming iterative training, and freedom from model-specific attention modulation. Specifically, we first propose a key hypothesize, \ie, homogeneous regions exhibit consistent discrepancy from logits to GT distributions, whereas heterogeneous regions manifest distinct discrepancy. The hypothesis enables the distribution discrepancy to directly characterize semantic information. Therefore, as illustrated in~\Cref{fig:intro}, our key idea lies in deriving an analytical solution for the distribution discrepancy to obtain the segmentation maps directly, bypassing the time-consuming iterative training (or model-specific attention modulation) required to solve the optimal logits prior to constructing the segmentation maps. In other words, we reformulate the optimization of the distribution discrepancy into the analytic solution of the distribution discrepancy.

In this work, we first validate the feasibility of the key hypothesis via exploration experiments and then address the reformulation from two distinct perspectives (optimal path and maximum velocity), ultimately achieving state-of-the-art performance on eight OVSS benchmark datasets. Our principal contributions are as follows:
\begin{itemize}[leftmargin=*]
\item We propose a key hypothesis, \ie, the distributional discrepancy between logits and GT effectively reveals semantic characteristics, and validate its feasibility.
\item We propose a straightforward method that directly solves the distribution discrepancy to characterize segmentation maps without logits-optimization.
\item Our method achieves state-of-the-art performance on eight OVSS benchmark datasets without requiring time-consuming training or model-specific modulation.
\end{itemize}

%% file: sec/2_relatedwork.tex
\begin{figure*}[t]
  \centering
    \includegraphics[width=1\linewidth]{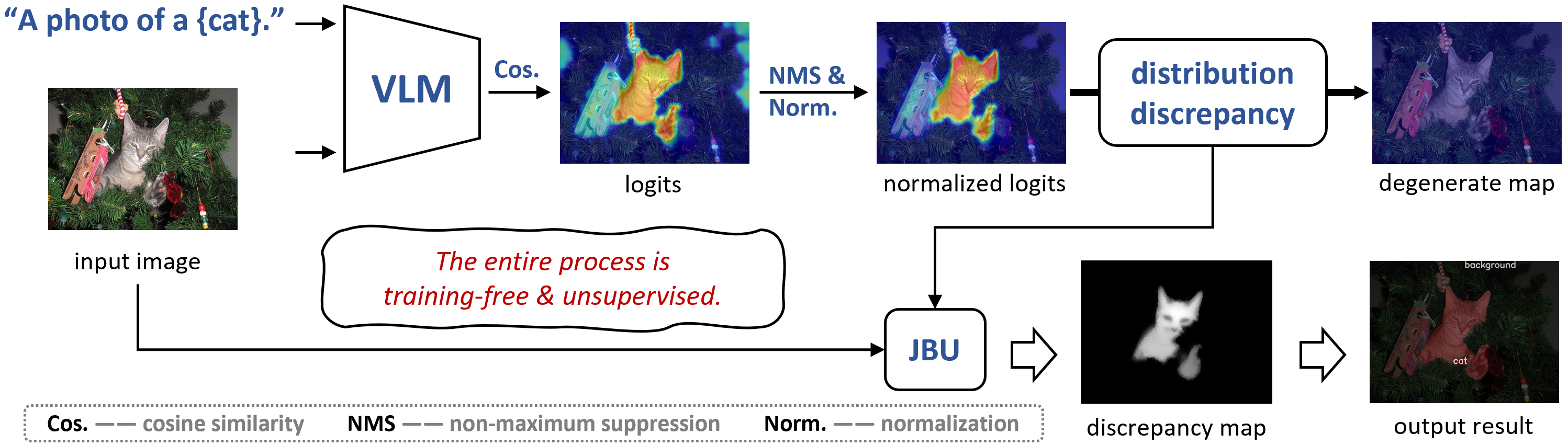}
    \caption{Overview of the proposed method. The pipeline begins by computing the cosine similarity (Cos.) between vision-language features to obtain logits, followed by non-maximum suppression (NMS) and normalization (Norm.) to filter low-confidence patches. The core step involves calculating the distribution discrepancy between the normalized logits and the degenerate distribution map. This solved discrepancy is then upsampled to the original image resolution via joint bilateral upsampling (JBU) to obtain the discrepancy map. Finally, the semantic map is generated by applying an $\arg\max$ operation to the discrepancy map.}
  \label{fig:overview}
\end{figure*}

\section{Related Works}
\subsection{Open-vocabulary semantic segmentation}
OVSS primarily encompasses two distinct methodologies: 1) transductive learning-based generative approaches~\cite{gu2020context,bucher2019zero,cheng2021sign,wu2023diffumask,wang2025diffusion}, and 2) inductive learning-driven discriminative methods~\cite{ding2022decoupling,zhou2022extract,zhou2023zegclip,xian2019semantic,shi2025llmformer}. Within transductive frameworks, generative techniques demand prior awareness of unseen categories in open-world environments. Given the theoretical inaccessibility of such priors, existing methods utilize open-vocabulary text embeddings to establish cross-modal connections between textual and visual domains, thereby generating these priors. These solutions~\cite{gu2020context,bucher2019zero,pastore2021closer} generate embeddings for novel categories by fusing visual embeddings with textual semantic representations derived from existing priors. Conversely, discriminative strategies utilize inductive mechanisms to deduce unseen categories via knowledge acquired during training, circumventing requirements for novel category priors. To acquire knowledge with potent representational capabilities, contemporary state-of-the-art systems predominantly adopt either knowledge distillation~\cite{ding2022decoupling,xu2022simple,yu2023zero,liang2023open} or feature adaptation approaches~\cite{zhou2023zegclip,kwon2023probabilistic,xu2023side}. Knowledge distillation integrates image-level discriminative capacities from vision-language models into mask-aware segmentation networks, whereas feature adaptation techniques directly adapt vision-language models (\eg, CLIP) as backbone architectures to convert image-level classification competencies into pixel-wise discriminative capabilities. Consequently, state-of-the-art research has focused on the adaptation mechanisms of CLIP for OVSS.

\subsection{CLIP adaptation for OVSS}

Existing CLIP adaptation approaches predominantly categorize into three paradigms: 1) Joint Fine-tuning~\cite{cho2024cat,jiao2023learning,li2024relationship}: This paradigm involves fine-tuning CLIP with segmentation-specific components to enhance dense prediction capabilities. For instance, CAT-Seg~\cite{cho2023cat} implements cost-based CLIP fine-tuning, while MAFT~\cite{jiao2023learning} leverages attention bias for classification-oriented refinement. 2) Pre Fine-tuning\cite{wu2023clipself,wu2024clim,xu2022groupvit}: This approach enhances CLIP's alignment granularity through fine-grained vision-language contrastive learning. Specifically, CLIM~\cite{wu2024clim} employs mosaic augmentation to create composite images for region-text contrastive learning, whereas CLIPSelf~\cite{wu2023clipself} maximizes cosine similarity between regional representations and corresponding crops. 3) Training-Free Adaptation\cite{lan2024clearclip,wang2024sclip,lan2024proxyclip,stojnic2025lposs}: This paradigm modulates CLIP's final residual attention layer or integrates vision foundation models (VFM)\cite{zhang2022dino,oquab2023dinov2} to boost alignment granularity. The omission of residual connections in CLIP's final layer significantly enhances visual embedding granularity\cite{lan2024clearclip}, motivating research into refined self-attention mechanisms for spatial alignment. This paradigm comprises two subcategories: a) VFM-proxy: This subcategory augments CLIP with VFMs' dense representations. ProxyCLIP~\cite{lan2024proxyclip} substitutes CLIP's self-attention with DINO's~\cite{zhang2022dino} visual self-similarity, and CASS~\cite{kim2025distilling} integrates DINO features via spectral graph distillation. b) Self-proxy: This subcategory constructs novel self-attention matrices from CLIP's internal embeddings. SCLIP~\cite{wang2024sclip} employs summed query-query and key-key attention matrices, while ClearCLIP~\cite{lan2024clearclip} and NACLIP~\cite{hajimiri2025pay} respectively utilize query-query and key-key matrices as replacements.

Ultimately, these methods fundamentally aim to optimize the discrepancy between the logits and GT distributions to generate optimal logits, which are subsequently converted into final segmentation maps. Departing from these conventional approaches, we avoid optimizing the distribution discrepancy and instead directly characterize segmentation maps by analytically solving for the discrepancy.

%% file: sec/3_approach.tex
\section{Approach}
\subsection{Problem definition} \label{Problem Definition}
Given an image ${\bm I}$ and class-specific textual descriptions $\{{\bm L}_i\}^{N_c}_{i=1}$, where ${\bm L_i}$ denotes the textual description of the $i$-th class and $N_c$ represents the total number of classes. OVSS assigns each pixel in ${\bm I}$ to the class label $i$ corresponding to the most semantically relevant $\bm{L}_i$ with $N_c$ being dynamic during inference.

\subsection{Overview} \label{Overview}
Current methods aim to optimize the logits --- defined as the cosine similarity between vision and language features --- by minimizing the distribution discrepancy between the logits and GT distributions, thereby seeking the optimal logits solutions:
\begin{align}
    \mathcal{Q}^* = \arg\min_{\mathcal{Q}} \mathbf{D}(\mathcal{P}\Vert \mathcal{Q}), \label{eq:op}\quad \bm{M} = \arg\max_{N_c}(\mathcal{Q}^*),
\end{align}
where $\mathcal{P}$ and $\mathcal{Q}$ denote the GT and the logits distributions, respectively, and $\mathbf{D}(\cdot)$ measures the distribution discrepancy, like the Kullback-Leibler (KL) divergence. The semantic maps $\bm{M}$ are then derived by the optimal logits. We hypothesize that patches from identical classes exhibit consistent distribution discrepancy, while those from different classes manifest significant divergence. Building upon the hypothesis, we can directly employ the distribution discrepancy $\mathbf{D}(\mathcal{P}\Vert \mathcal{Q})$ to obtain the semantic maps without logits-optimization process, thus reformulating the optimization formulation in Equation~\ref{eq:op} into solving an analytic solution for the distribution discrepancy. However, since the GT distribution $\mathcal{P}$ is unavailable during inference, our idea is to replace the GT distribution $\mathcal{P}$ with an equivalent surrogate distribution. This enables quantifying the distribution discrepancy between the logits and this surrogate distribution to derive the semantic map:
\vspace{-0.15cm}
\begin{align}
    \bm{M} = \arg\max_{N_c} \mathbf{D}(\mathcal{S}\Vert \mathcal{Q}), \label{eq:as}
\end{align}
where $\mathcal{S}$ denotes the surrogate distribution.

Therefore, in this work, we focus on two key technical challenges: (1) which distribution to choose as the surrogate distribution for GT; and (2) how to derive the analytic solution for the distribution discrepancy. For the first challenge, we opt for a degenerate distribution as the surrogate; for the second challenge, we approach it from two distinct perspectives (optimal path and maximum velocity). The overview of our method is illustrated in~\Cref{fig:overview}, depicting our pipeline that first computes the cosine similarity (Cos.) between visual-language features to obtain the logits, and applies non-maximum suppression (NMS) and normalization (Norm.) to obtain the normalized logits. Subsequently, we solve the distribution discrepancy from the logits distribution to a degenerate distribution. Finally, the solved distribution discrepancy undergoes joint bilateral upsampling (JBU) to generate the discrepancy map, with the final segmentation result derived through an $\arg\max$ operation. Compared to the existing logits-optimization methods seeking optimal solutions, we reformulate to an analytic solution form, achieving independence from GT annotations, elimination of time-consuming training, and freedom from model-specific modulation.

\subsection{Reformulation analysis} \label{reformulate}
The reformulation from the logits-optimization formulation in Equation~\ref{eq:op} to the analytic solution in Equation~\ref{eq:as} relies on the hypothesis that distribution discrepancy effectively captures semantic information and that replacing the degenerate distribution with the GT distribution is valid. Consequently, this section aims to analyze the feasibility of characterizing semantic maps through distribution discrepancy and to investigate the replacement of the degenerate distribution with the GT distribution.

\begin{figure}[t]
  \centering
    \includegraphics[width=\linewidth]{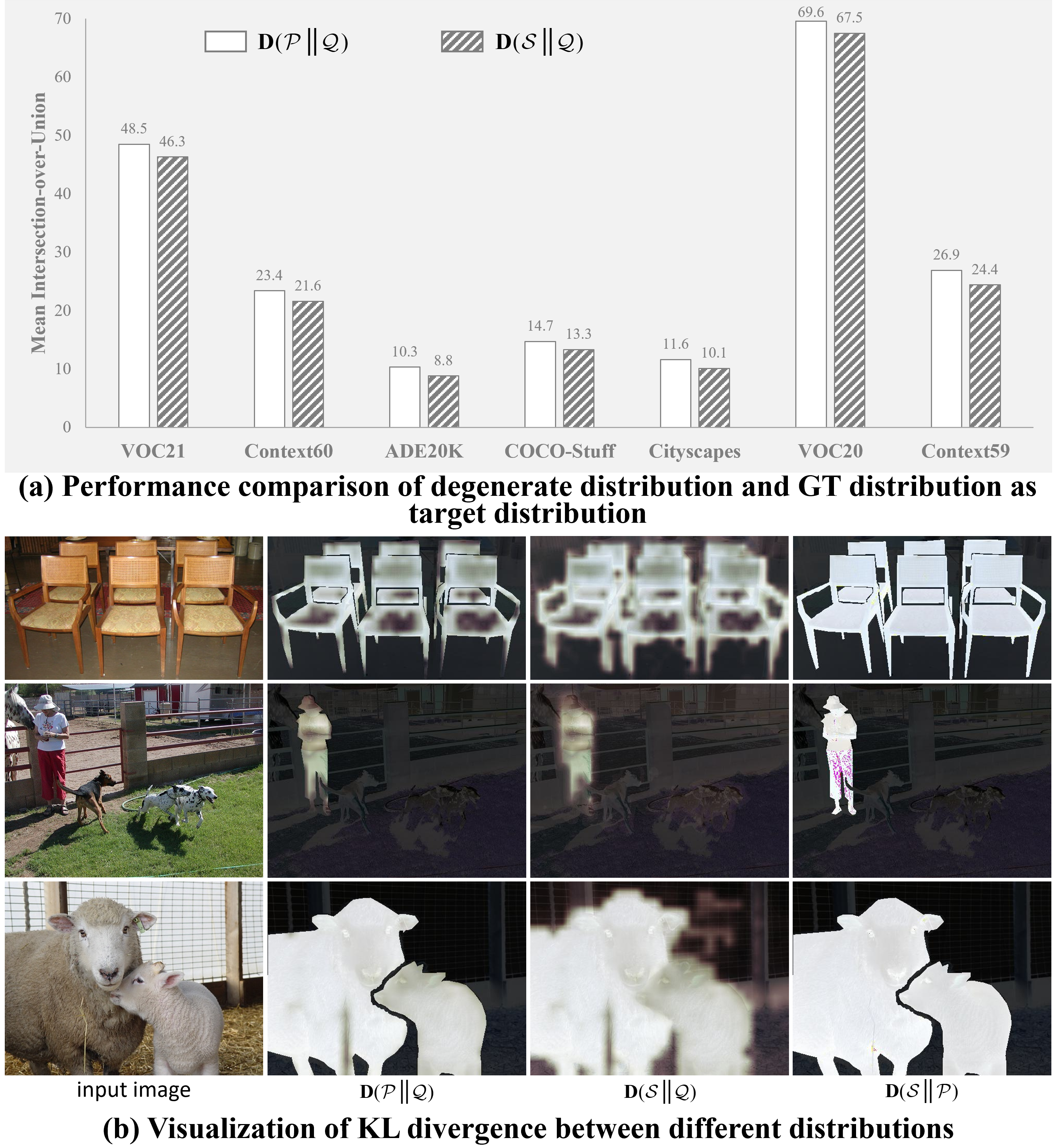}
  \caption{Reformulation analysis results. (a) Performance comparison targeting the degenerate distribution $\mathcal{S}$ and the GT distribution $\mathcal{P}$ demonstrates that the equivalence of the two distributions. (b) KL divergence visualization across three transmission scenarios confirms distribution discrepancy's capacity to capture semantic information.}
  \label{fig:exp}
\end{figure}

Specifically, we employ the KL divergence to quantify distribution discrepancy $\mathbf{D}(\cdot)$ and examine three transmission scenarios: 1) from logits to GT distribution (\ie, $\mathbf{D}(\mathcal{P}\Vert \mathcal{Q})$), 2) from logits to degenerate distribution (\ie, $\mathbf{D}(\mathcal{S}\Vert \mathcal{Q})$), and 3) from GT to degenerate distribution (\ie, $\mathbf{D}(\mathcal{S}\Vert \mathcal{P})$). The semantic maps are generated by applying an $\arg\max$ operation to these KL divergence measurements. As shown in~\Cref{fig:exp} (a), we present a quantitative comparison of semantic maps derived from $\mathbf{D}(\mathcal{P}\Vert \mathcal{Q})$ and $\mathbf{D}(\mathcal{S}\Vert \mathcal{Q})$ across five benchmark datasets. Experimental results demonstrate highly consistent quantitative performance for both scenarios across all datasets, indicating that the solution spaces for targeting the degenerate and GT distributions coincide. This confirms the feasibility of substituting the GT distribution with the degenerate distribution for direct solving the distribution discrepancy. Furthermore, visualization results for $\mathbf{D}(\mathcal{P}\Vert \mathcal{Q})$ and $\mathbf{D}(\mathcal{S}\Vert \mathcal{Q})$ in~\Cref{fig:exp} (b) indicate that regions of the identical classes exhibit highly consistent distribution discrepancy, confirming our hypothesis regarding the consistency of distribution discrepancy across homogeneous regions and demonstrating the ability of the distribution discrepancy to capture semantic information. For $\mathbf{D}(\mathcal{S}\Vert \mathcal{P})$, the distribution discrepancy visualization in~\Cref{fig:exp} (b) near-complete overlap with the GT distribution, implying that $\mathcal{S}$ and $\mathcal{P}$ occupy antipodal positions in the feature space. While logits-optimization approaches optimize towards the GT endpoint, our methodology computes the distribution discrepancy towards the degenerate endpoint.

In summary, although the KL divergency between the logits and the degenerate distribution can capture the semantic information, its quantitative performance remains limited. Therefore, the core challenge involves accurately measuring the distribution discrepancy from the logits to the degenerate distribution. We approach this challenge from two distinct perspectives: 1) solving the optimal transport path to quantify the distribution discrepancy, and 2) solving the maximum transport velocity to define this discrepancy.

\subsection{Optimal path} \label{Optimal Transport}
In this section, we focus on solving the optimal transport path between the logits distribution and the degenerate distribution. Intuitively, for regions of the same category, the degradation paths should exhibit consistency. Consequently, the optimal transport path can be leveraged to quantify the distribution discrepancy. To this end, we formulate the problem of measuring the distribution discrepancy as an optimal transport task, specifically by seeking the optimal transformation from the logits to the degenerate distribution, which corresponds to the optimal path between them.

Given the normalized logits $\bm{f}^c \in \mathbb{R}^N$ for the $c$-th category and the degenerate map $\bm{f}^t \in \mathbb{R}^N$, where $N$ denotes the number of patches, $\bm{f}^t$ follows a degenerate (uniform) distribution, \ie, $\bm{f}^t=\frac{1}{N}\mathbf{1}_{N}$. Under Sinkhorn's theorem~\cite{cuturi2013sinkhorn}, the problem of solving the optimal transport path $\bm{p}^c \in \mathbb{R}^{N}$ between $\bm{f}^c$ and $\bm{f}^t$ is formulated as:
\begin{align}
    \bm{p}^c&=\operatorname{Norm.}(\sum_i\bm{\pi}^*_{i,j}\cdot\bm{C}_{i,j}), \label{eq:path}\\ 
    \bm{\pi}^*&=\min_{\bm{\pi}} \sum_{i,j} \bm{C}_{i,j}\bm{\pi}_{i,j} - \epsilon\sum_{i,j}\bm{\pi}_{i,j}(\ln\bm{\pi}_{i,j}-1), \label{eq:ot}
\end{align}
where $\bm{\pi}^*\in \mathbb{R}^{N\times N}$ represents the optimal transport matrix from $\bm{f}^c$ to $\bm{f}^t$, and $\epsilon$ is regularization scalar. $\bm{C}\in \mathbb{R}^{N\times N}$ denotes the cost matrix between $\bm{f}^c$ and $\bm{f}^t$, which can be formulated as layer-wise averaged self-attention tensor:
\begin{align}
    \bm{C}=\sum_{b,h}w_{b,h}\bm{S}_{b,h},
\end{align}
where $\bm{S}_{b,h}$ and $w_{b,h}$ denote the self-attention tensor and scalar weight for the $h$-th head in the $b$-th transformer block, respectively. To calculate the optimal transport path $\bm{p}^c$ via Equation~\ref{eq:path}, we first solve Equation~\ref{eq:ot} to obtain $\bm{\pi}^*$. Applying Lagrange Multiplier method, we can directly solve the unique solution of Equation~\ref{eq:ot}, which is as:
\begin{align}
    \bm{\pi}^*=\operatorname{diag}(\bm{\mu})\bm{K}\operatorname{diag}(\bm{\nu}), \label{eq:lag}
\end{align}
where the Gibbs kernel matrix $\bm{K}=\exp(-\bm{C}/\epsilon)$, and $\bm{\mu}\in \mathbb{R}^N$, $\bm{\nu}\in \mathbb{R}^N$ denote two unknown scaling parameters. According to the law of conservation of mass, these two unknown parameters are subject to the following constraints:
\begin{align}
    \bm{\mu}\odot(\bm{K}\bm{\nu})=\bm{f}^c, \quad \bm{\nu} \odot(\bm{K}^\top \bm{\mu})=\bm{f}^t,
\end{align}
where $\odot$ denotes element-wise product. And utilizing the Sinkhorn iteration algorithm, we update $\bm{\mu}$ and $\bm{\nu}$ alternately:
\begin{align}
    \bm{\mu}^{(l+1)}\leftarrow\frac{\bm{f}^c}{\bm{K}\bm{\nu}^{(l)}}, \quad \bm{\nu}^{(l+1)}\leftarrow\frac{\bm{f}^t}{\bm{K}^{\top}\bm{\mu}^{(l+1)}}, \label{eq:it}
\end{align}
where $l$ denotes the iteration index, and $\bm{\nu}^{(0)}=\mathbf{1}_N$. Therefore, the optimal transport path $\bm{p}^c$ can be solved according to Equation~\ref{eq:path}, ~\ref{eq:lag}, and~\ref{eq:it}. Finally, we reshape $\bm{p}^c$ to the downsampled size $h\times w$ with $N=h\cdot w$, and then upscale it via JBU to the original image size $H\times W$ to obtain the discrepancy map $\bm{m}^c$. Thus, the output result $\bm{M}$ is derived as:
\vspace{-0.2cm}
\begin{align}
    \bm{M}=\arg\max_{c}(\{\bm{m}^c\}^{N_c}_{c=1}),\quad c=1,2,...,N_c.\label{eq:out}
\end{align}

\subsection{Maximum velocity} \label{Markov Chain}
Intuitively, transport velocity plays a critical role in quantifying distribution discrepancy: for identical transport paths, reduced velocity prolongs transport duration, thereby amplifying the overall discrepancy. In this section, we focus on determining the transport velocity for the transitions from the logits to the degenerate distribution. Given that the degenerate distribution constitutes a special case of the stationary distribution, we formulate the velocity-determination task as a Markov process problem, specifically by assessing the Markov process through which the logits converges to the stationary distribution.

Inspired by~\cite{karmann2025repurposing}, given a transition matrix $\bm{T}\in \mathbb{R}^{N\times N}$, the Markov process can be defined as:
\begin{align}
    {\bm{f}^c}^{(l)}={\bm{f}^c}^{(0)}\cdot \bm{T}^l, \quad l=1,2,...
\end{align}
where $l$ denotes the Markov chain step. Since layer-wise averaged self-attention tensors naturally capture inter-patch transition probabilities, they serve as candidates for $\bm{T}$. However, Markov processes require $\bm{T}$ to be a strictly positive and irreducible, aperiodic stochastic matrix. To satisfy these conditions, we employ iterative proportional fitting (IPF) to transform the candidate self-attention tensor into a doubly stochastic form:
\begin{align}
    \bm{T}^{(k-\frac{1}{2})}_{i,j} \leftarrow \frac{\bm{T}^{(k-1)}_{i,j}}{\sum_u\bm{T}^{(k-1)}_{u,j}}, \quad \bm{T}^{(k+1)}_{i,j} \leftarrow \frac{\bm{T}^{(k-\frac{1}{2})}_{i,j}}{\sum_v\bm{T}^{(k-\frac{1}{2})}_{i,v}}, 
\end{align}
where $k\in \mathbb{Z}^+$ represents the iterations, initialized with $\bm{T}^{(0)}=\sum_{b,h}w_{b,h}\bm{S}_{b,h}$. Typically, $\bm{T}$ satisfies the Markov requirements after $15$ iterations. As $l\rightarrow +\infty$, repeated application of $\bm{T}$ drives ${\bm{f}^c}^{(l)}$ component-wise toward the stationary distribution, \ie, ${\bm{f}^c}^{(+\infty)}=\frac{1}{N}\mathbf{1}_N$. Given that patches of identical categories should converge to the stationary distribution at equivalent Markov steps, we define the maximum transport velocity $\bm{v}^c \in \mathbb{R}^N$ for each patch as the reciprocal of its convergence steps:
\begin{align}
    \bm{v}^c_i=\max\{\frac{1}{l},\,l\in \mathbb{Z}^+| \left |{\bm{f}^c_i}^{(l)} -{\bm{f}^c_i}^{(l-1)}\right | \leq \tau\}.\label{eq:ve}
\end{align}
Equation~\ref{eq:ve} indicates that when the probability variation of patch $i$ between consecutive steps no longer changes with the Markov process, in other words, patch $i$'s probability variation falls below threshold $\tau$, \ie, $\left |{\bm{f}^c_i}^{(l)} -{\bm{f}^c_i}^{(l-1)}\right | \leq \tau$, its convergence step $l$ is identified; the reciprocal $\frac{1}{l}$ then quantifies its maximum transport velocity. Finally, employing the same upsampling scheme to $\bm{v}^c$ as in Section~\ref{Optimal Transport} yields the discrepancy map $\bm{m}^c$, with the segmentation results computed via Equation~\ref{eq:out}.

%% file: sec/4_experiments.tex
\input{tables/system_comp}

\section{Experiments} \label{Experiments}

\subsection{Experiment Settings}
\noindent \textbf{Datasets \& Metric.} Following the unsupervised training-free paradigm, we conduct evaluations on eight standard benchmark datasets: 1) the Pascal VOC2012 series~\cite{pascal-voc-2012} comprises two benchmarks, \ie, VOC21 with $21$ categories and VOC20 with $20$ categories, derived from VOC21 by excluding the background category; 2) the Pascal Context series~\cite{mottaghi_cvpr14} comprises two benchmarks, \ie, Context60 with $60$ categories and Context59 with $59$ categories, derived from Context60 by excluding the background category; 3) the COCO series~\cite{caesar2018cvpr} comprises COCO-Stuff with $171$ categories and COCO-Obj with $81$ categories; 4) ADE20K~\cite{zhou2019semantic} contains $150$ categories; and 5) Cityscapes~\cite{cordts2016cityscapes} contains $19$ categories for autonomous driving. All experiments are evaluated using the mean Intersection-over-Union (mIoU).

\noindent \textbf{Implementation.} All experiments are conducted using PyTorch with MMSegmentation and diffusers. CLIP models with ViT-B/16 and ViT-L/14 architectures are employed to construct the logits. Inspired by diffusion-based segmentation methods~\cite{tian2024diffuse,karmann2025repurposing}, we utilize Stable Diffusion~\cite{rombach2022high}, particularly version 2 (SD2), to extract the self-attention tensor. The model weights are sourced from the Hugging Face transformers package. Noise-free latent features of images are directly encoded, and a single-step unconditional denoising process is performed to extract self-attention tensors from each block. To ensure computational efficiency, experiments are conducted using 16-bit floating-point precision. The regularization scalar $\epsilon$ is set to $0.1$. The number of iterations for Equation~\ref{eq:it} is set to 50. The parameter $\tau$ is set to $0.3$. Existing approaches predominantly leverage sliding-window inference to enhance performance, whereas our method employs direct whole-image inference without requiring any post-processing.

\begin{figure*}[t]
  \centering
    \includegraphics[width=1\linewidth]{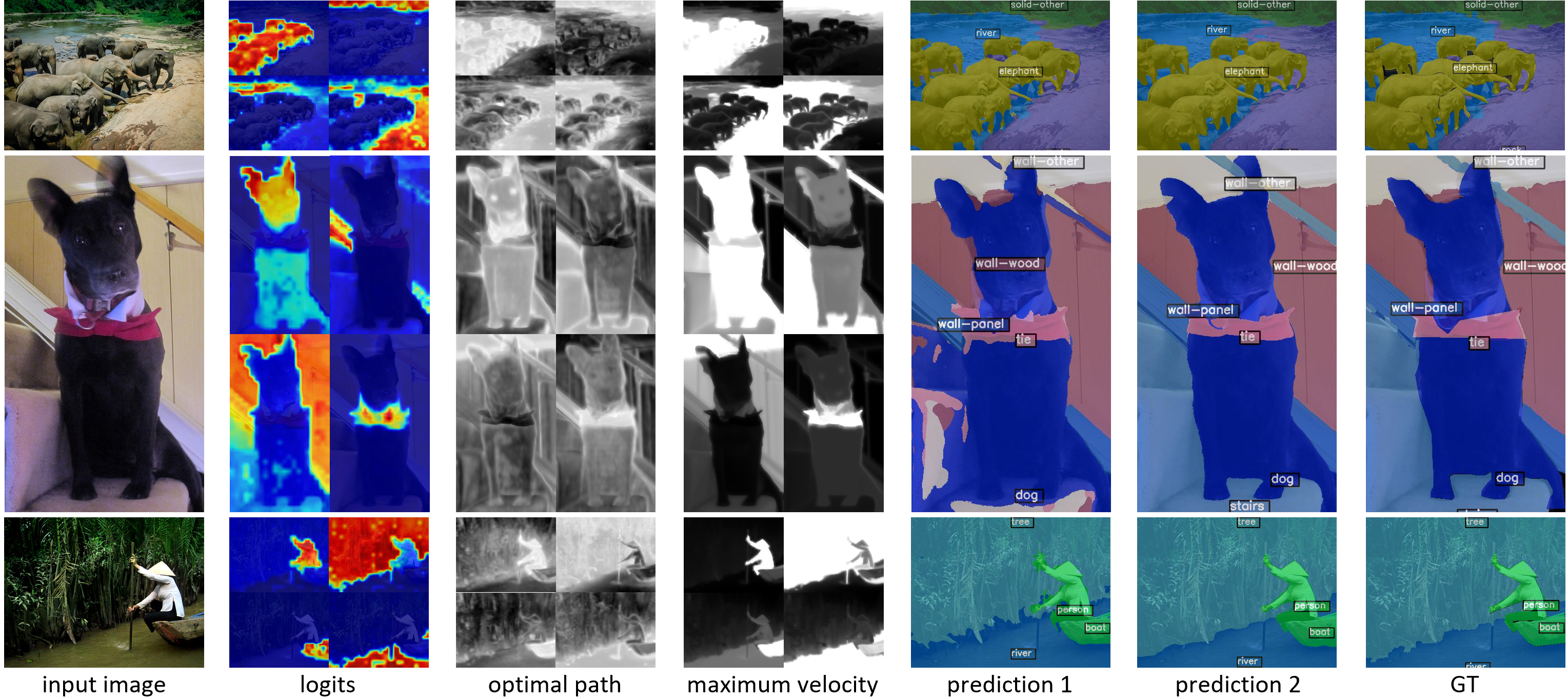}
        \caption{Visualization of outputs from each stage of our pipeline. We present the logits and distribution discrepancies for each category across the input image, segmentation maps, and GT. Predictions $1$ and $2$ represent the segmentation maps obtained by solving the optimal path and maximum velocity formulations, respectively. Visualization of both distribution discrepancies demonstrates that the optimal path mode is more sensitive to high-frequency regions than the maximum velocity mode. For instance, the visualization results from the optimal path mode in the second image can clearly depict the internal contours of the dog.}
  \label{fig:vis1}
\end{figure*}

\subsection{System level comparison}
We conduct quantitative evaluations on eight benchmark datasets to compare against existing OVSS methods. To ensure experimental fairness, we perform evaluations on models with two distinct scales: the CLIP base and large models. Existing methods involve time-consuming iterative training and model-specific attention modulation paradigms. As shown in Table~\ref{tab:system_comp}, the performance comparison across all benchmarks demonstrates that: (1) Our method achieves an average improvement of approximately $2$ mIoU points on all benchmarks under both base and large scales. For instance, under the base scale, the optimal path and maximum velocity modes achieve mIoU improvements of $+1.8\%$ and $+2.5\%$ mIoU over CASS~\cite{kim2025distilling}, respectively. (2) Our approach consistently ranks in the top two positions on nearly every benchmark. Specifically, our method achieves state-of-the-art performance on VOC21, Context60, VOC20, COCO-Stuff, and Cityscapes. (3) The maximum velocity mode exhibits marginal superiority over the optimal path mode, yielding mIoU gains of $+0.7\%$ and $+0.6\%$ under the base and large scale, respectively.

\Cref{fig:vis1} displays visualizations of the outputs from each stage of our pipeline, including the normalized logits, distribution discrepancies for both optimal path and maximum velocity modes, and the final segmentation maps. Comparisons of these visualizations reveal two key observations. First, the optimal path mode exhibits higher sensitivity to intra-class high-frequency textures, whereas the maximum velocity mode demonstrates stronger responsiveness to inter-class distinctions. Second, as a result, the optimal path mode occasionally fails to segment large background regions with inconsistent illumination (\eg, the staircase in the second image, which is mis-segmented into incorrect regions), whereas the maximum velocity mode precisely delineates inter-class boundaries.

\subsection{Ablation study}
\noindent \textbf{Component analysis.} Table~\ref{tab:component_ablation} presents the quantitative analysis of each component's contribution in the proposed method. We establish a baseline model (Row (I)) that utilizes the CLIP base model to directly output raw vision-language features and compute their cosine similarity, termed raw logits; the final segmentation maps are obtained through an $\arg\max$ operation and bilinear upsampling. Row (II) enhances Row (I) by computing the KL divergence map between the raw logits and the degenerate map, then applying the prior segmentation map acquisition process to this KL divergence map. Building upon Row (II), Row (III) incorporates NMS on the raw logits to mask low-confidence patches. Row (IV) further improves Row (III) by replacing bilinear upsampling with JBU. Rows (V) and (VI) modify Row (IV) by substituting the KL divergence map with the proposed optimal path map and maximum velocity map, respectively. Finally, Row (VII) augments Row (IV) through fusion of the optimal path map and maximum velocity map. The experimental results from component-wise performance comparisons reveal the following insights: (1) Constructing distribution discrepancy yields performance improvements comparable to optimizing logits; specifically, Row (II) achieves a significant average increase of $+8.9\%$ in mIoU over the baseline. (2) The two proposed modes for constructing distribution discrepancy demonstrate exceptional performance, with respective gains of $+22.2\%$ and $+23.0\%$ in mIoU. (3) However, fusing the two distinct distribution discrepancy maps leads to performance degradation.

\begin{figure*}[t]
  \centering
    \includegraphics[width=1\linewidth]{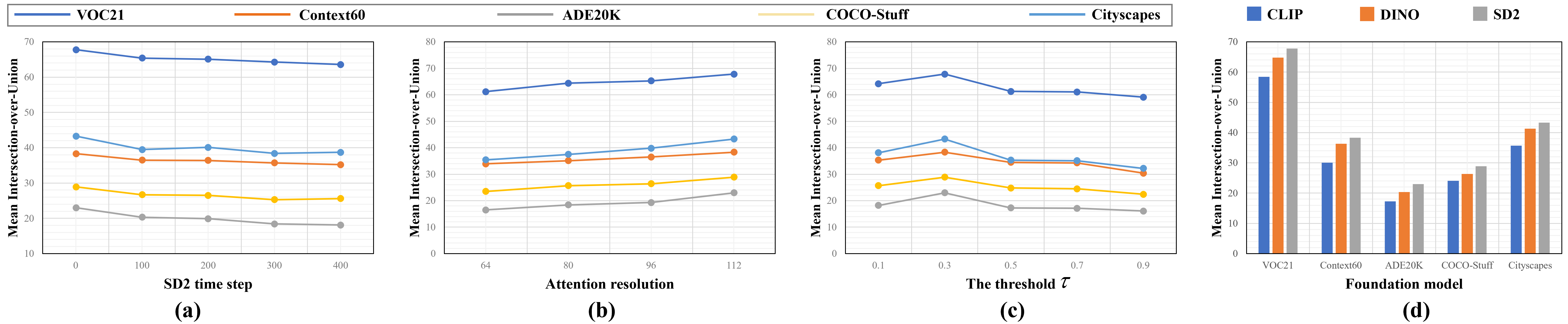}
    \caption{Quantitative evaluation of SD2 time step, attention resolution, threshold $\tau$ and foundation model on standard benchmarks is presented, with each metric represented by a distinct color (unit: \%). Here, SD2 time step denotes the denoising step in SD2, and attention resolution refers to the input size of SD2. All comparisons were performed using the maximum velocity mode.}
  \label{fig:exp1}
\end{figure*}

\noindent \textbf{Time step analysis.} The number of denoising steps in diffusion models influences feature representations during encoding; while increasing these steps typically enhances generation quality in image synthesis tasks, its impact on constructing the maximum velocity map requires specific investigation. To quantify this effect, we conducted comparative experiments across varying step configurations. As illustrated in~\Cref{fig:exp1} (a), quantitative evaluation across five benchmark datasets demonstrates that fewer denoising steps yield superior performance. We hypothesize that this phenomenon stems from the encoding process avoiding noise injection, thereby ensuring deterministic feature extraction without addressing generative variability.

\input{tables/component_ablation}

\noindent \textbf{Attention analysis.} Self-attention tensors play a pivotal role in the proposed method's pipeline; for instance, during the construction of the maximum velocity map, these tensors are utilized to formulate the transition matrix of Markov processes, which governs the generation of the maximum velocity map. In this evaluation, we quantitatively compare self-attention tensors derived from diverse foundation models, aggregation strategies across blocks in SD2, and the impact of resolution variations. The results demonstrate: (1) as shown in~\Cref{fig:exp1} (d), SD2's self-attention tensors outperform those from three ViT-based foundation models across five benchmark datasets for transition matrix construction; (2) as presented in Table~\ref{tab:variant_comp}, aggregating SD2 self-attention tensors reveals that combining $\text{up}_0$ and $\text{up}_1$ blocks achieves optimal performance among six kinds of strategies; (3) as illustrated in~\Cref{fig:exp1} (b), higher resolutions consistently yield superior performance.

\noindent \textbf{Threshold $\tau$ analysis.} During the construction of the maximum velocity map, threshold $\tau$ determines the convergence time as the logits distribution approaches the degenerate distribution; theoretically, higher thresholds accelerate convergence. This part examines the impact of threshold variations: as shown in~\Cref{fig:vis1} (c), performance peaks at $\tau=0.3$ and decreases monotonically with increasing threshold. We hypothesize that higher thresholds cause premature degeneration, preventing the logits distribution from reaching the optimal degradation state.

\noindent \textbf{Logits-optimization analysis.} Numerous methods directly modify the CLIP attention mechanism to optimize logits. We focus on comparing two mainstream approaches: transforming the original query-key computation into query-query mode and key-key mode, respectively. Quantitative comparisons in Table~\ref{tab:variant_comp} demonstrate that these schemes yield improvements of $2.9\%$ and $4.6\%$ respectively compared to the original attention mechanism.

\input{tables/variant_comp}

%% file: tables/system_comp.tex
\begingroup
\renewcommand{\arraystretch}{1.0}
\begin{table*}[t]
  \centering
  \caption{{Quantitative evaluation on standard benchmarks (unit: \%).} \textbf{$\dagger$}: Re-implementation with OpenAI's weights. \textbf{$\ddagger$}: Re-implementation with its DINO-B/16 variant. \textbf{M.M.} denotes model-specific modulation. \textbf{I.T.} denotes iterative training. Here, the best results are shown in bold and the second-best results are underlined.}
  \setlength{\tabcolsep}{0.4em}
  \resizebox{\textwidth}{!}{
  \begin{tabular}{lcc|ccccccccc}
    \toprule
    \textbf{{Model}} & {\textbf{Logits Model}} & {\textbf{Paradigm}} & {\textbf{VOC21}} & {\textbf{Context60}} & {\textbf{COCO-Obj}} & {\textbf{VOC20}} & {\textbf{Context59}} & {\textbf{COCO-Stuff}} & {\textbf{Cityscapes}} & {\textbf{ADE20K}} & {\textbf{Avg.}} \\
    \midrule
    \midrule
    FreeDa~\cite{barsellotti2024training} & CLIP ViT-B/16 & M.M. & - & - & - & 85.6 & {43.1} & {27.8} & 36.7 & {22.4} & -\\
    CLIP~\cite{radford2021learning} & CLIP ViT-B/16 & I.T. & 18.6 & 7.8 & 6.5 & 49.1 & 11.2 & 7.2 & 6.7 & 3.2 & 13.8\\
    ReCo~\cite{shin2022reco} & CLIP ViT-B/16 & M.M. & 25.1 & 19.9 & 15.7 & 57.7 & 22.3 & 14.8 & 21.6 & 11.2 & 23.5\\
    MaskCLIP~\cite{zhou2022extract} & CLIP ViT-B/16 & I.T. & 38.3 & 23.6 & 20.6 & 74.9 & 26.4 & 16.4 & 12.6 & 9.8 & 27.9\\
    GroupViT~\cite{xu2022groupvit} & CLIP ViT-B/16 & I.T. & 50.4 & 18.7 & 27.5 & 79.7 & 23.4 & 15.3 & 11.1 & 9.2 & 29.4\\
    CLIPtrase~\cite{shao2024explore} & CLIP ViT-B/16 & M.M. & 50.9 & 29.9 & \textbf{43.6} & 81.0 & 33.8 & 22.8 & 21.3 & 16.4 & 32.7\\
    TCL~\cite{cha2023learning} & CLIP ViT-B/16 & I.T. & 55.0 & 30.4 & 31.6 & 83.2 & 33.9 & 22.4 & 24.0 & 17.1 & 37.2\\
    CLIPSurgery~\cite{li2023clip} & CLIP ViT-B/16 & I.T. & {55.2} & {30.3} & {29.7} & {77.5} & {33.4} & {22.2} & {33.1} &{16.1}&{37.2} \\
    LaVG~\cite{kang2024defense} & CLIP ViT-B/16 & M.M. & {62.1} & 31.6 & 34.2 & {82.5} & 34.7 & 23.2 & 26.2 & 15.8 & 38.8\\
    GEM~\cite{bousselham2024grounding} & CLIP ViT-B/16 & I.T. & 58.7 & 32.0 & 32.9 & 81.7 & 35.6 & 23.9 & 32.6 & 16.9 & 39.3\\
    CaR~\cite{sun2024clip} & CLIP ViT-B/16 & M.M. & 48.6 & 13.6 & 15.4 & 73.7 & 18.4 & - & - & 5.4 & -\\
    ClearCLIP~\cite{lan2024clearclip} & CLIP ViT-B/16 & M.M. & 51.8 & 32.6 & 33.0 & 80.9 & 35.9 & 23.9 & 30.0 & 16.7 & 38.1\\
    SCLIP~\cite{wang2024sclip} & CLIP ViT-B/16 & M.M. & 59.1 & 30.4 & 30.5 & 80.4 & 34.1 & 22.4 & 32.2 & 16.1 & 38.2\\
    NACLIP~\cite{hajimiri2025pay} & CLIP ViT-B/16 & M.M. & 58.9 & 32.2 & 33.2 & 79.7 & 35.2 & 23.3 & 35.5 & 17.4 & 39.4\\
    CLIP-DINOiser$^\dagger$~\cite{wysoczanska2024clip} & CLIP ViT-B/16 & I.T. & 62.1 & 32.4 & 34.8 & 80.9 & 35.9 & 24.6 & 31.7 & 20.0 & 40.3\\
    ProxyCLIP$^\ddagger$~\cite{lan2024proxyclip} & CLIP ViT-B/16 & M.M. & 59.1 & {35.2} & 36.2 & 78.2 & {38.8} & {26.2} & {38.1} & {19.6} & {41.4}\\
    SC-CLIP~\cite{bai2024self} & CLIP ViT-B/16 & M.M. & {64.6} & {36.8} & {37.7} & {84.3} & {40.1} & {26.6} & {41.0} &{20.1} & {43.9} \\
    CASS~\cite{kim2025distilling} & CLIP ViT-B/16 & M.M. & {65.8} & {36.7} & {37.8} & {{87.8}} & {40.2} & {26.7} & {39.4} &{20.4} & {44.4} \\
    \rowcolor{blue!5}
    \textbf{{Ours}}  &  &  & &  &  &  &  &  &  &  & \\
    \rowcolor{blue!5}
    \textit{{-\space solve optimal path}}  & CLIP ViT-B/16 & & \underline{66.9} & \underline{37.6} & 38.9 & \underline{88.6} & \underline{44.4} & \underline{28.6} & \underline{41.7} & \underline{22.8} & \underline{46.2}\\
    \rowcolor{blue!5}
    \textit{{-\space solve maximum velocity}}  & CLIP ViT-B/16 & & \textbf{67.8} & \textbf{38.3} & \underline{39.6} & \textbf{89.0} & \textbf{44.9} & \textbf{28.9} & \textbf{43.3} & \textbf{23.0} & \textbf{46.9}\\
    \midrule
    CLIP~\cite{radford2021learning} & CLIP ViT-L/14 & I.T. & 10.3 & 4.5 & 4.4 & 19.9 & 5.7 & 3.2 & 3.2 & 1.9 & 6.6\\
    MaskCLIP~\cite{zhou2022extract} & CLIP ViT-L/14 & I.T. & 24.8 & 9.7 & 10.2 & 30.1 & 13.0 & 9.0 & 12.1 & 7.1 & 14.5\\
    SCLIP~\cite{wang2024sclip} & CLIP ViT-L/14 & M.M.  & 44.4 & 22.3 & 24.9 & 70.6 & 25.2 & 16.5 & 21.3 & 10.9 & 29.5\\
    GEM~\cite{bousselham2024grounding} & CLIP ViT-L/14 & I.T. & 45.2 & 25.5 & 28.3 & 83.7 & 28.1 & 19.2 & 27.1 & 13.2 & 33.8\\
    CLIPSurgery~\cite{li2023clip} & CLIP ViT-L/14 & I.T. & {47.9} & {27.3} & {28.1} & {84.3} & {31.0} & {21.4} & {29.7} &{17.3}&{35.9} \\
    PnP-OVSS~\cite{luo2024emergent} &  CLIP ViT-L/14 & M.M. & - & - & 36.2 & 51.3 & 28.0 & 17.9 & - & 14.2 & -\\
    NACLIP~\cite{hajimiri2025pay} & CLIP ViT-L/14 & M.M. & 52.1 & 28.7 & 29.9 & 78.6 & 32.1 & 21.4 & 31.4 & 17.3 & 36.4\\
    ClearCLIP~\cite{lan2024clearclip} & CLIP ViT-L/14 & M.M. & 48.6 & 28.0 & 28.6 & 84.8 & 31.5 & 21.2 & 32.1 & 16.9 & 36.5\\
    ProxyCLIP$^\ddagger$~\cite{lan2024proxyclip} & CLIP ViT-L/14 & M.M. & 58.1 & 34.1 & 37.4 & 82.0 & 37.3 & 25.5 & 38.1 & 21.2 & 41.7\\
    SC-CLIP~\cite{bai2024self} & CLIP ViT-L/14 & M.M. & {65.0} & {36.9} & {40.5} & {88.3} & {40.6} & {26.9} & {41.3} & {21.7} & {45.2} \\
    \rowcolor{blue!5}
    \textbf{{Ours}}  &  & &  &  &  &  &  &  &  &  & \\
    \rowcolor{blue!5}
    \textit{{-\space solve optimal path}}  & CLIP ViT-L/14 & & \underline{68.2} & \underline{37.9} & \underline{42.3} & \underline{89.7} & \underline{44.8} & \underline{28.9} & \underline{42.3} &\underline{23.4} &\underline{47.2}\\
    \rowcolor{blue!5}
    \textit{{-\space solve maximum velocity}}  & CLIP ViT-L/14 & & \textbf{68.9} & \textbf{38.7} & \textbf{42.9} & \textbf{90.1} & \textbf{45.3} & \textbf{29.2} & \textbf{43.9} & \underline{23.4} & \textbf{47.8}\\
\bottomrule
  \end{tabular}
  }
  \label{tab:system_comp}
\end{table*}
\endgroup

%% file: tables/component_ablation.tex
\begingroup
\renewcommand{\arraystretch}{1.0}
\begin{table}[t]
\centering
\caption{{Effect of different components (unit: \%).} \textbf{Baseline}: apply $\arg\max$ to the raw logits without optimization or analytical solution. \textbf{KL}: Kullback-Leibler (KL) divergence. \textbf{O.P.}: solving the optimal path. \textbf{M.V.}: solving the maximum velocity.}
\label{tab:component_ablation}
\resizebox{\linewidth}{!}{
    \begin{tabular}{ll!{\vrule height 10pt}ccccc}
    \toprule
    &\textbf{Components} & \textbf{VOC21} & \textbf{COCO-Stuff} & \textbf{Cityscapes} & \textbf{ADE20K} & \textbf{Avg.}\\
    \midrule\midrule
    \textbf{(I)} & Baseline & 18.6 & 7.2 & 6.7 & 3.2 & 8.9 \\
    \midrule
    \textbf{(II)} & \textbf{(I)} + KL   & 44.2 & 12.1 & 8.6 & 6.4 & 17.8\\
    \textbf{(III)} & \textbf{(II)} + NMS   & 45.9 & 13.0 & 9.6 & 7.7 & 19.1\\
    \textbf{(IV)} &\textbf{(III)} + JBU  & 46.3 & 13.3 & 10.1 & 8.8 & 19.6\\
    \textbf{(V)} &\textbf{(IV)} + O.P. & 66.9 & 28.6 & 41.7 & 22.8 & 40.0 \\
    \rowcolor{blue!5}
    \textbf{(VI)} &\textbf{(IV)} + M.V.  & 67.8  & 28.9 & 43.3 & 23.0 & 40.8\\
    \textbf{(VII)} &\textbf{(V)} + \textbf{(VI)}  & 64.9 & 26.8 & 41.4 & 20.5 & 38.4\\
    \bottomrule
    \end{tabular}
    }
\end{table}
\endgroup

%% file: tables/variant_comp.tex
\begingroup
\renewcommand{\arraystretch}{1.0}
\begin{table}[t]
\centering
\caption{Effect of different strategy for our components (unit: \%). \textbf{q-q mean}: the average attention tensor between queries across all layers. \textbf{k-k mean}: the average attention tensor between keys across all layers. All comparisons were performed using the maximum velocity mode.}
\label{tab:variant_comp}
\resizebox{\linewidth}{!}{
    \begin{tabular}{ll!{\vrule height 10pt}ccccc}
    \toprule
    & \textbf{Variants} & \textbf{VOC21} & \textbf{COCO-Stuff} & \textbf{Cityscapes} & \textbf{ADE20K} & \textbf{Avg.}\\
    \midrule
    \midrule
    \multicolumn{7}{c}{self-attention weight combination ($\text{down}_0,\text{down}_1,\text{up}_0,\text{up}_1,\text{up}_2$)}\\
    \midrule
    \textbf{(I)} & ($1,0,0,0,0$) & 62.1 & 24.6 & 37.5 & 17.6 & 35.5\\
    \textbf{(II)} & ($0,1,0,0,0$) & 63.3 & 26.1 & 40.1 & 18.5 & 37.0 \\
    \textbf{(III)} & ($0,0,1,0,0$) & 65.0 & 26.7 & 41.9& 19.6 & 38.3 \\
    \textbf{(IV)} & ($0,0,0,1,0$) & 66.2 & 27.8 & 42.8 & 21.8 & 39.7 \\
    \textbf{(V)} & ($0,0,0,0,1$) & 65.3 & 26.9 & 42.4& 20.2 & 38.7 \\
    \rowcolor{blue!5}
    \textbf{(VI)} & ($0,0,0.5,0.5,0$) & 67.8  & 28.9 & 43.3 & 23.0 & 40.8\\
    \midrule
    \multicolumn{7}{c}{training-free logits-optimization methods}\\
    \midrule
    \textbf{(I)} & origin & 63.7 & 26.2 & 37.5 & 17.5 & 36.2 \\
    \textbf{(II)} & q-q mean & 65.5 & 27.3 & 42.1 & 21.5 & 39.1 \\
    \rowcolor{blue!5}
    \textbf{(III)} & k-k mean & 67.8  & 28.9 & 43.3 & 23.0 & 40.8\\
    \bottomrule
    \end{tabular}}
\end{table}
\endgroup

%% file: sec/5_conclusion.tex
\section{Conclusion}
In this work, we propose a novel approach that bypasses the logits-optimization paradigm, which typically employs time-consuming iterative training or model-specific attention modulation to minimize the discrepancy between logits and GT distributions. Our approach directly calculates the distribution discrepancy to quantify segmentation maps. The efficacy of our method stems from a hypothesis foundation: the distribution discrepancy effectively captures semantic information. We first validate this hypothesis and then introduce a degenerate distribution as a surrogate for GT. And then, we devise two distinct modes for discrepancy construction: an optimal path map via optimal transport theory, and a maximum velocity map based on Markov processes. Comprehensive evaluations across eight benchmark datasets demonstrate the effectiveness of our method. 


%% file: sec/6_acknowledgments.tex
\section*{Acknowledgments}

This work was supported in part by National Natural Science Foundation of China under Grant 62306165, 62376233, in part by Science and Technology on Sonar Laboratory under grant 2024-JCJQ-LB-32/07, in part by State Grid Corporation Headquarters Technology Project 52120025004V-463-FGS, in part by Fundamental Research Funds for the Central Universities under Grant 20720250031, in part by Xiaomi Young Talents Program award.

%% file: sec/X_suppl.tex
\clearpage
\setcounter{page}{1}
\maketitlesupplementary

\section{Proof}
Given the cost matrix $\bm{C}\in\mathbb{R}^{N\times N}$ and the regularization scalar $\epsilon$, the objective is to solve the following equation:
\begin{align}
    \bm{\pi}^*=\min_{\bm{\pi}} \sum_{i,j} \bm{C}_{i,j}\bm{\pi}_{i,j} - \epsilon\sum_{i,j}\bm{\pi}_{i,j}(\ln\bm{\pi}_{i,j}-1),
\end{align}
subject to marginal constraints:
\begin{align}
    \sum_j \bm{\pi}_{i,j}=\bm{f}^c_i, \sum_i \bm{\pi}_{i,j}=\bm{f}^t_j, \quad \forall i,j, \quad \bm{\pi}_{i,j} \geq 0,
\end{align}
where $\sum_i\bm{f}^c_i=1,\sum_j\bm{f}^t_j=1$. By introducing Lagrange multipliers $\alpha\in \mathbb{R}^N$ and $\beta\in \mathbb{R}^N$, the Lagrangian is defined as:
\begin{equation}
\begin{split}
    &\mathcal{L}(\bm{\pi},\alpha,\beta)=\min_{\bm{\pi}} \sum_{i,j} \bm{C}_{i,j}\bm{\pi}_{i,j} - \epsilon\sum_{i,j}\bm{\pi}_{i,j}(\ln\bm{\pi}_{i,j}-1)\\
    &+\sum_i\alpha_i(\bm{f}^c_i-\sum_j\bm{\pi}_{i,j})+\sum_j\beta_j(\bm{f}^t_j-\sum_i\bm{\pi}_{i,j}).
\end{split}
\end{equation}
Next, taking the partial derivative of $\mathcal{L}$ with respect to $\bm{\pi}_{i,j}$ yields:
\begin{align}
    \frac{\partial \mathcal{L}}{\partial\bm{\pi}_{i,j}}=\bm{C}_{i,j}-\epsilon\ln\bm{\pi}_{i,j}-\alpha_i-\beta_j=0.
\end{align}
Solving this equation for $\bm{\pi}_{i,j}$:
\begin{equation}
\begin{split}
    \bm{\pi}_{i,j}&=\exp(-\frac{\bm{C}_{i,j}-\alpha_i-\beta_j}{\epsilon}-1)\\
    &=\exp(-\frac{\bm{C}_{i,j}}{\epsilon})\exp(\frac{\alpha_i}{\epsilon})\exp(\frac{\beta_j}{\epsilon})\cdot e^{-1}.
\end{split}
\end{equation}
We define:
\begin{align}
    \bm{\mu}_i=\exp(\frac{\alpha_i}{\epsilon}-\frac{1}{2}),\quad \bm{\nu}_j=\exp(\frac{\beta_j}{\epsilon}-\frac{1}{2}).
\end{align}
Then, the solution is expressed as:
\begin{align}
    \bm{\pi}_{i,j}=\bm{\mu}_{i}\bm{K}_{i,j}\bm{\nu}_{j}.
\end{align}
Substitute the above solution into the marginal constraints yields:
\begin{equation}
\begin{split}
    \sum_j\bm{\pi}_{i,j}&=\bm{\mu}_i\sum_j\bm{K}_{i,j}\bm{\nu}_j=\bm{f}^c_i,\\
    \sum_i\bm{\pi}_{i,j}&=\bm{\nu}_j\sum_i\bm{\mu}_i\bm{K}_{i,j}=\bm{f}^t_j.
\end{split}
\end{equation}
Thus,
\begin{equation}
\begin{split}
    \bm{\mu}_i=\frac{\bm{f}^c_i}{\sum_j\bm{K}_{i,j}\bm{\nu}_j},\quad \bm{\nu}_j=\frac{\bm{f}^t_j}{\sum_i\bm{\mu}_i\bm{K}_{i,j}}.
\end{split}
\end{equation}
In summary, this yields the Sinkhorn iteration format:
\begin{equation}
\begin{split}
    \bm{\mu}^{(l+1)}=\frac{\bm{f}^c}{\bm{K}\bm{\nu}^{(l)}},\quad \bm{\nu}^{(l+1)}=\frac{\bm{f}^t}{\bm{K}^{\top}\bm{\mu}^{(l+1)}}.
\end{split}
\end{equation}

\begin{algorithm}[tb]
\caption{Our pipeline}
\label{alg:algorithm}
\textbf{Input}: Image $\bm{I}$, class-specific textual descriptions $\{\bm{L}_i\}^{N_c}_{i=1}$\\
\textbf{Parameter}: degenerate map $f^t$, optimal path (or maximum velocity) operation ${\bf D}(\cdot)$\\
\textbf{Output}: the segmentation map $M$
\begin{algorithmic}[1] 
\STATE $f=\text{Cos.}(f^I, f^L)$ \COMMENT{Construct the logits.}
\STATE $N_c=\text{unique}(\arg\max(f))$ \COMMENT{Execute category early rejection to obtain the most probable category $N_c$.}
\STATE $f=\text{Norm.}(\text{NMS}(f))$ \COMMENT{Apply NMS and normalization to the logits.}
\FOR{$c$ to $N_c$}
\STATE $m^c={\bf D}(f^c,f^t)$ \COMMENT{Execute the proposed method.}
\ENDFOR
\STATE $M=\arg\max_{c}(\{m^c\}^{N_c}_{c=1})$ \COMMENT{Construct the segmentation map}
\RETURN $M$
\end{algorithmic}
\end{algorithm}

\begin{figure*}[t]
  \centering
    \includegraphics[width=1\linewidth]{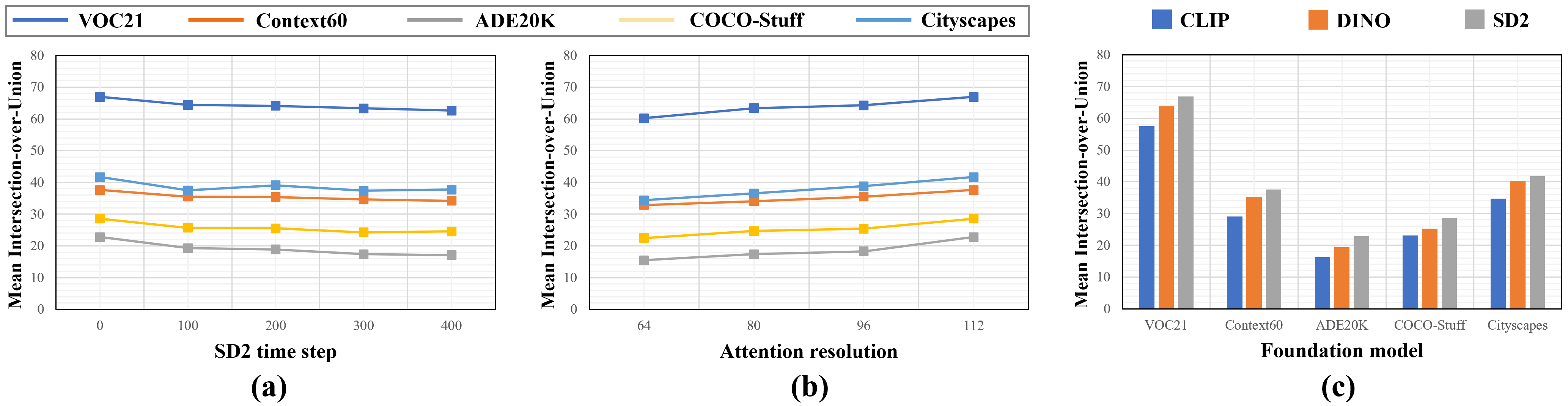}
    \caption{Quantitative evaluation of SD2 time step, attention resolution, and foundation model on standard benchmarks is presented, with each metric represented by a distinct color (unit: \%). Here, SD2 time step denotes the denoising step in SD2, and attention resolution refers to the input size of SD2. All comparisons were performed using the optimal path mode.}
  \label{fig:sup-exp1}
\end{figure*}

\section{More Details}
\paragraph{Non-maximum suppression.} Given that existing visual language models are constrained by coarse-grained multimodal training paradigms, the resulting logits often contain numerous misaligned patches, which serve as noise and interfere with downstream fine-grained tasks. In this work, this noise disrupts the distribution transmission process, particularly for similar patches, resulting in consistent differences in distribution between noisy and clean patches. Consequently, noise removal is crucial. Although numerous methods have made significant contributions, most concentrate on noise removal while preserving clean logit regions, necessitating precise localization of noise regions. We propose to treat low-confidence patches as noise and set their values to $-\infty$, thereby ensuring a probability distribution of zero in the softmax-normalized output. Specifically, we define patches with confidence less than $0.9$ as noise. This means that it is possible to establish the distribution discrepancy representing semantic information by relying only on a small subset of reliable logits distribution.
\paragraph{Normalization.} We adopt $\mathbf{softmax}$ operation to obtain the normalized logits.
\paragraph{Joint bilateral upsampling.} Joint Bilateral Upsampling (JBU) is an edge-preserving image upsampling technique that integrates spatial and range information. It is widely employed in computer vision to align low-resolution processing results, such as segmentation maps and depth maps, with high-resolution reference images, including the original input. The core principle of JBU involves leveraging the texture structure of the high-resolution reference image to guide the upsampling of low-resolution features, thereby preventing edge blurring. The JBU process is defined as:
\begin{align}
    D^H(p)=\frac{1}{k_p}\sum_{q\in\Omega}D^L(q)\cdot f(\Vert p-q \Vert)\cdot g(\Vert I^H(p)-I^H(q) \Vert),
\end{align}
where $D^L$ and $I^H$ denote low-resolution segmentation maps and high-resolution RGB image. $D^H$ denote the upsampled results. $p$ denotes pixel location in a high-resolution image. $q$ denotes pixel positions in the neighborhood $\Omega$ centered at $p$. $k_p=\sum_{q\in\Omega}f(\Vert p-q \Vert)\cdot g(\Vert I^H(p)-I^H(q) \Vert)$. $f(\cdot)$ and $g(\cdot)$ denote space gaussian kernel and range gaussian kernel. Usually, $f(x)=e^{-\frac{x^2}{\sigma^2_s}}$, and $g(x)=e^{-\frac{x^2}{\sigma^2_r}}$. We set $\sigma^2_s=1,\sigma^2_r=0.1$.
\paragraph{Category early rejection.} When the number of categories is large, inference time increases substantially. To mitigate this issue while leveraging the observation that most images contain only a few semantic categories, we apply the $\arg\max$ operation to the normalized logits to identify the most probable category before executing the proposed method.

\paragraph{Pseudo algorithm.} The algorithm is illustrated in Algorithm~\ref{alg:algorithm}.

\input{tables/sup_variant_comp}

\section{More results}
\paragraph{Ablation about optimal path.} We conduct component ablation experiments under the optimal path mode. As illustrated in~\Cref{fig:sup-exp1}, our analysis reveals that the effect of denoising step length confirms that single-step denoising generates deterministic self-attention tensors with optimal performance. Moreover, higher resolution of the attention tensor generally correlates with improved performance. Consequently, SD2 was selected as the self-attention tensor extraction model to achieve the best results. Table~\ref{tab:sup-variant_comp} demonstrates the self-attention weight combination and training-free logits-optimization strategy under this mode, with results consistent with the maximum velocity mode.

\paragraph{Computational complexity analysis.} To verify the efficiency of our method, we conduct efficiency analysis on VOC21 benchmark using an NVIDIA RTX 3090 GPU. The experiments were performed under default settings: input resolution of $512\times512$ pixels, SD2 time step set to $0$, and the logits model scale set to base/16. As shown
\begin{wraptable}{r}{0.6\linewidth}
	\centering
	\vspace{-0.5cm}
    \hspace{-0.8cm}
	\resizebox{\linewidth}{!}{
      \begin{tabular}{l!{\vrule height 10pt}cccc}
        \hline
        Model & FLOPs(G)$\downarrow$ & Params(M)$\downarrow$ & Inference time(sec.)$\downarrow$  & mIoU(\%)$\uparrow$ \\ \hline
        CLIP & 52.2  & 149.6 & 0.08 & 18.6\\
        ProxyCLIP & 103.2  & 235.4 & 0.16 & 59.1 \\
        CASS & 1675.5  & 265.7 & 3.11 & 65.8 \\
        \rowcolor{blue!5}
        \textbf{Ours (O.P.)} & 351.3  & 1006.8 & 0.59{\scriptsize \textcolor{red}{(-2.52)}} & 66.9{\scriptsize \textcolor{red}{(+1.1)}} \\
        \rowcolor{blue!5}
        \textbf{Ours (M.V.)} & 339.2  & 1006.8 & 0.57{\scriptsize \textcolor{red}{(-2.54)}} & 67.8{\scriptsize \textcolor{red}{(+2.0)}} \\ \hline
        \multicolumn{5}{c}{the inference time for each component}\\ \hline
        Model & logits time(sec.)$\downarrow$ & attention time(sec.)$\downarrow$ & distribution time(sec.)$\downarrow$ & JBU(sec.)$\downarrow$ \\ \hline
        \rowcolor{blue!5}
        \textbf{Ours (O.P.)} & 0.08 & 0.25 & 0.10 & 0.10 \\
        \rowcolor{blue!5}
        \textbf{Ours (M.V.)} & 0.08 & 0.25 & 0.12 & 0.10 \\ \hline
      \end{tabular}
  }
\end{wraptable}
in the table (please zoom in for details), our method achieves an optimal balance between performance and computational efficiency. For example, compared to CASS, our approach delivers superior performance while maintaining faster inference speed. In addition, we conduct a detailed breakdown of the inference time for each component. The ``Logits" and ``Attention" times represent the computational costs for generating logits via CLIP and attention maps via SD2, respectively. ``Distribution" refers to the processing time of our proposed method (including Optimal Path (O.P.) \& Maximum Velocity (M.V.)), while ``JBU" denotes the time required for upsampling operation. Our analysis reveals that the primary bottleneck in inference time stems from generating attention maps using SD2. Even when setting the time step to 0 without introducing noise and performing only the intermediate attention map calculation, the overall inference speed remains constrained by this component. In contrast, O.P.\& M.V. introduce minimal computational overhead, requiring only approximately 0.1 seconds of processing time (50 iterations for O.P. and $\tau=0.3$ for M.V. can converge quickly).

\paragraph{VFM integration for structural/spatial priors.} Since VFMs provide good self-attention (spatial prior), its integration into OVSS has become common practice. For existing methods (including ours), integrating high/low-quality self-attention of VFMs inevitably increases/degrades performance. However, our approach demonstrates two advantages based on high- and low-quality self-attention. \ding{172} Integrating low-quality self-attention, our method demonstrates excellent robustness—even integrating the original CLIP’s self-attention
\begin{wraptable}{r}{0.6\linewidth}
	\centering
    \hspace{-0.8cm}
	\resizebox{\linewidth}{!}{
      \begin{tabular}{l!{\vrule height 10pt}ccccc}
        \hline
        Model & VOC21 & Context60 & ADE20K & COCO-Stuff & Cityscapes \\ \hline
        original CLIP-B/16 & 18.6  & 7.8 & 3.2 & 7.2 & 6.7 \\
        \rowcolor{blue!5}
        integrating self-attention via maximum velocity &   &  &  &  &  \\
        \rowcolor{blue!5}
        \textit{{-\space CLIP-B/16}} & 58.5{\scriptsize \textcolor{red}{(+39.9)}}  & 30.1{\scriptsize \textcolor{red}{(+22.3)}} & 17.3{\scriptsize \textcolor{red}{(+14.1)}} & 24.1{\scriptsize \textcolor{red}{(+16.9)}} & 35.7{\scriptsize \textcolor{red}{(+29.0)}} \\
        \rowcolor{blue!5}
        \textit{{-\space DINO-B/8}} & 64.8  & 36.3 & 20.4 & 26.3 & 41.3 \\
        \rowcolor{blue!5}
        \textit{{-\space DINOv2-B/14}} & 66.9  & 36.9 & 21.3 & 27.9 & 42.7 \\
        \rowcolor{blue!5}
        \textit{{-\space DINOv2-B/14 w/ Registers}} & \textbf{69.8}  & \textit{38.0} & \textit{22.3} & \textbf{29.3} & \textbf{44.6} \\
        \rowcolor{blue!5}
        \textit{{-\space SD2}} & \textit{67.8}  & \textbf{38.3} & \textbf{23.0} & \textit{28.9} & \textit{43.3} \\ \hline
      \end{tabular}
  }
\end{wraptable}
still achieves high performance. While the original CLIP suffers severe performance degradation due to its low-quality self-attention, our method achieves an average improvement of 24.4 points, demonstrating its excellent robustness to lower-quality self-attention. \ding{173} Several SOTA methods generate high-quality self-attention and achieve excellent performance. By integrating these approaches, our method achieves further improvements, demonstrating its ability to fully leverage high-quality self-attention. In addition, we observe that integrating DINOv2 with Registers yields stronger performance improvements due to its higher-quality self-attention. Therefore, while a more powerful VFM would undoubtedly improve performance (due to its higher-quality spatial priors), we emphasize that our method does not rely on model-specific attention improvement—enabling flexible integration of diverse VFMs.

\paragraph{NMS analysis.} With a higher NMS threshold, we obtain
\begin{wrapfigure}{r}{0.4\linewidth}
    \centering
    \vspace{-0.4cm}
    \hspace{-0.8cm}
    \includegraphics[width=1\linewidth]{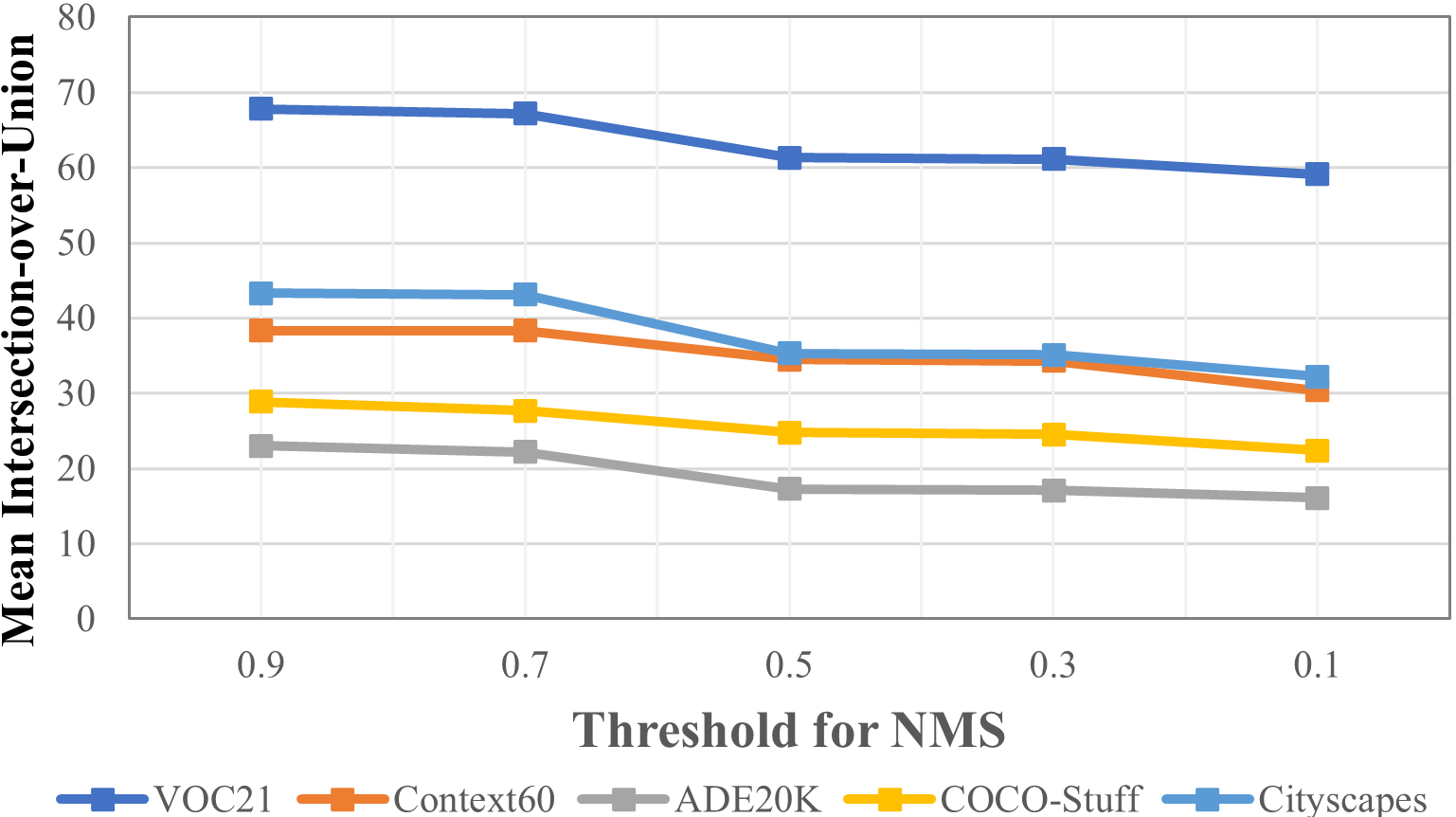}
    \vspace{-0.6cm}
\end{wrapfigure}
more reliable logits patches despite fewer filtered patches, without damaging performance. Conversely, lowering the threshold to get more patches (not necessarily more reliable) impairs performance, as evidenced by the significant drop below 0.5 threshold in the figure below (please zoom in for details).

\paragraph{Versatility analysis.} Our method can indeed be regarded as a flexible and general post-processing step that operates independently of any specific setup. As demonstrated in the 
\begin{wraptable}{r}{0.5\linewidth}
	\centering
	\vspace{-0.5cm}
    \hspace{-0.8cm}
	\resizebox{1\linewidth}{!}{
		\begin{tabular}{cc|cccccccc}
			\hline
			Model  & Scale  & VOC21  & Context60 & COCO-Obj & VOC20 & Context59 & COCO-Stuff & Cityscapes & ADE20K \\ \hline
            CASS    & B/16 & 65.8 & 36.7 & 37.8 & 87.8 & 40.2 & 26.7 & 39.4 & 20.4 \\
            \rowcolor{blue!5}
            w/ O.P. &      & 68.2{\scriptsize \textcolor{red}{(+2.4)}} & 38.7{\scriptsize \textcolor{red}{(+2.0)}} & 40.1{\scriptsize \textcolor{red}{(+2.3)}} & 89.6{\scriptsize \textcolor{red}{(+1.8)}} & 45.5{\scriptsize \textcolor{red}{(+5.3)}} & 29.4{\scriptsize \textcolor{red}{(+2.7)}} & 42.8{\scriptsize \textcolor{red}{(+3.4)}} & 22.9{\scriptsize \textcolor{red}{(+2.5)}} \\
            \rowcolor{blue!5}
            w/ M.V. &      & 69.5{\scriptsize \textcolor{red}{(+3.7)}} & 39.1{\scriptsize \textcolor{red}{(+2.4)}} & 40.6{\scriptsize \textcolor{red}{(+2.8)}} & 90.0{\scriptsize \textcolor{red}{(+2.2)}} & 45.9{\scriptsize \textcolor{red}{(+5.7)}} & 29.8{\scriptsize \textcolor{red}{(+3.1)}} & 44.4{\scriptsize \textcolor{red}{(+5.0)}} & 23.1{\scriptsize \textcolor{red}{(+2.7)}} \\
            SC-CLIP & B/16 & 64.6 & 36.8 & 37.7 & 84.3 & 40.1 & 26.6 & 41.0 & 20.1 \\
            \rowcolor{blue!5}
            w/ O.P. &      & 68.1{\scriptsize \textcolor{red}{(+3.5)}} & 38.7{\scriptsize \textcolor{red}{(+1.9)}} & 40.0{\scriptsize \textcolor{red}{(+2.3)}} & 89.2{\scriptsize \textcolor{red}{(+4.9)}} & 45.3{\scriptsize \textcolor{red}{(+5.2)}} & 29.3{\scriptsize \textcolor{red}{(+2.7)}} & 42.9{\scriptsize \textcolor{red}{(+1.9)}} & 22.8{\scriptsize \textcolor{red}{(+2.7)}} \\
            \rowcolor{blue!5}
            w/ M.V. &      & 69.3{\scriptsize \textcolor{red}{(+4.7)}} & 39.1{\scriptsize \textcolor{red}{(+2.3)}} & 40.6{\scriptsize \textcolor{red}{(+2.9)}} & 89.6{\scriptsize \textcolor{red}{(+5.3)}} & 45.7{\scriptsize \textcolor{red}{(+5.6)}} & 29.8{\scriptsize \textcolor{red}{(+3.2)}} & 44.6{\scriptsize \textcolor{red}{(+3.6)}} & 23.0{\scriptsize \textcolor{red}{(+2.9)}} \\
            RF-CLIP & B/16 & 64.8 & 36.4 & 37.9 & 87.0 & 39.8 & 26.3 & 41.3 & 20.4 \\
            \rowcolor{blue!5}
            w/ O.P. &      & 67.8{\scriptsize \textcolor{red}{(+3.0)}} & 38.3{\scriptsize \textcolor{red}{(+1.9)}} & 40.1{\scriptsize \textcolor{red}{(+2.2)}} & 89.5{\scriptsize \textcolor{red}{(+2.5)}} & 44.7{\scriptsize \textcolor{red}{(+4.9)}} & 29.3{\scriptsize \textcolor{red}{(+3.0)}} & 42.9{\scriptsize \textcolor{red}{(+1.6)}} & 22.9{\scriptsize \textcolor{red}{(+2.5)}} \\
            \rowcolor{blue!5}
            w/ M.V. &      & 69.1{\scriptsize \textcolor{red}{(+4.3)}} & 39.0{\scriptsize \textcolor{red}{(+2.6)}} & 40.6{\scriptsize \textcolor{red}{(+2.7)}} & 90.0{\scriptsize \textcolor{red}{(+3.0)}} & 45.6{\scriptsize \textcolor{red}{(+5.8)}} & 29.7{\scriptsize \textcolor{red}{(+3.4)}} & 44.4{\scriptsize \textcolor{red}{(+3.1)}} & 23.1{\scriptsize \textcolor{red}{(+2.7)}} \\
            SC-CLIP & L/14 & 65.0 & 36.9 & 40.5 & 88.3 & 40.6 & 26.9 & 41.3 & 21.7 \\
            \rowcolor{blue!5}
            w/ O.P. &      & 69.3{\scriptsize \textcolor{red}{(+4.3)}} & 39.0{\scriptsize \textcolor{red}{(+2.1)}} & 42.7{\scriptsize \textcolor{red}{(+2.2)}} & 91.5{\scriptsize \textcolor{red}{(+3.2)}} & 45.8{\scriptsize \textcolor{red}{(+5.2)}} & 29.7{\scriptsize \textcolor{red}{(+2.8)}} & 43.4{\scriptsize \textcolor{red}{(+2.1)}} & 23.5{\scriptsize \textcolor{red}{(+1.8)}} \\
            \rowcolor{blue!5}
            w/ M.V. &      & 70.0{\scriptsize \textcolor{red}{(+5.0)}} & 39.8{\scriptsize \textcolor{red}{(+2.9)}} & 43.2{\scriptsize \textcolor{red}{(+2.7)}} & 91.9{\scriptsize \textcolor{red}{(+3.6)}} & 46.2{\scriptsize \textcolor{red}{(+5.6)}} & 30.0{\scriptsize \textcolor{red}{(+3.1)}} & 45.0{\scriptsize \textcolor{red}{(+3.7)}} & 23.6{\scriptsize \textcolor{red}{(+1.9)}} \\
            RF-CLIP & L/14 & 65.8 & 36.7 & 41.8 & 89.1 & 40.2 & 26.7 & 41.4 & 22.4 \\
            \rowcolor{blue!5}
            w/ O.P. &      & 69.5{\scriptsize \textcolor{red}{(+3.7)}} & 38.7{\scriptsize \textcolor{red}{(+2.0)}} & 43.6{\scriptsize \textcolor{red}{(+1.8)}} & 91.6{\scriptsize \textcolor{red}{(+2.5)}} & 45.5{\scriptsize \textcolor{red}{(+5.3)}} & 29.5{\scriptsize \textcolor{red}{(+2.8)}} & 43.4{\scriptsize \textcolor{red}{(+2.0)}} & 24.3{\scriptsize \textcolor{red}{(+1.9)}} \\
            \rowcolor{blue!5}
            w/ M.V. &      & 70.1{\scriptsize \textcolor{red}{(+4.3)}} & 39.5{\scriptsize \textcolor{red}{(+2.8)}} & 44.5{\scriptsize \textcolor{red}{(+2.7)}} & 92.0{\scriptsize \textcolor{red}{(+2.9)}} & 46.0{\scriptsize \textcolor{red}{(+5.8)}} & 30.0{\scriptsize \textcolor{red}{(+3.3)}} & 45.0{\scriptsize \textcolor{red}{(+3.6)}} & 24.6{\scriptsize \textcolor{red}{(+2.2)}} \\ \hline
		\end{tabular}
	}
	\vspace{-0.6cm}
\end{wraptable} 
table (please zoom in for details), we apply our approach to several SOTA methods, including RF-CLIP, CASS, and SC-CLIP. The experimental results show that integrating our method leads to significant performance improvements across existing SOTA approaches. We observe an approximate 5-point improvement on the Context59 benchmark, along with an average gain of 3 points across other benchmark datasets. These consistent improvements demonstrate that our method functions as a universal enhancement method for open-vocabulary tasks.

\paragraph{Discussion about Label Propagation Over Patches and Pixels for Open vocabulary Semantic Segmentation\cite{stojnic2025lposs}.} Paper\cite{stojnic2025lposs} essentially proposes a method for refining labels, which still follows the idea of logit-optimization, continuously pushing the logit towards the true label distribution. However, we don't care about the true label distribution; we only care about the difference between the degenerate distribution and the logit distribution, thus eliminating the need for logit-optimization. Therefore, our proposed method is fundamentally different from paper\cite{stojnic2025lposs}.

\paragraph{More visualization.} See the following figures.

\begin{figure*}[t]
  \centering
    \includegraphics[width=0.85\linewidth]{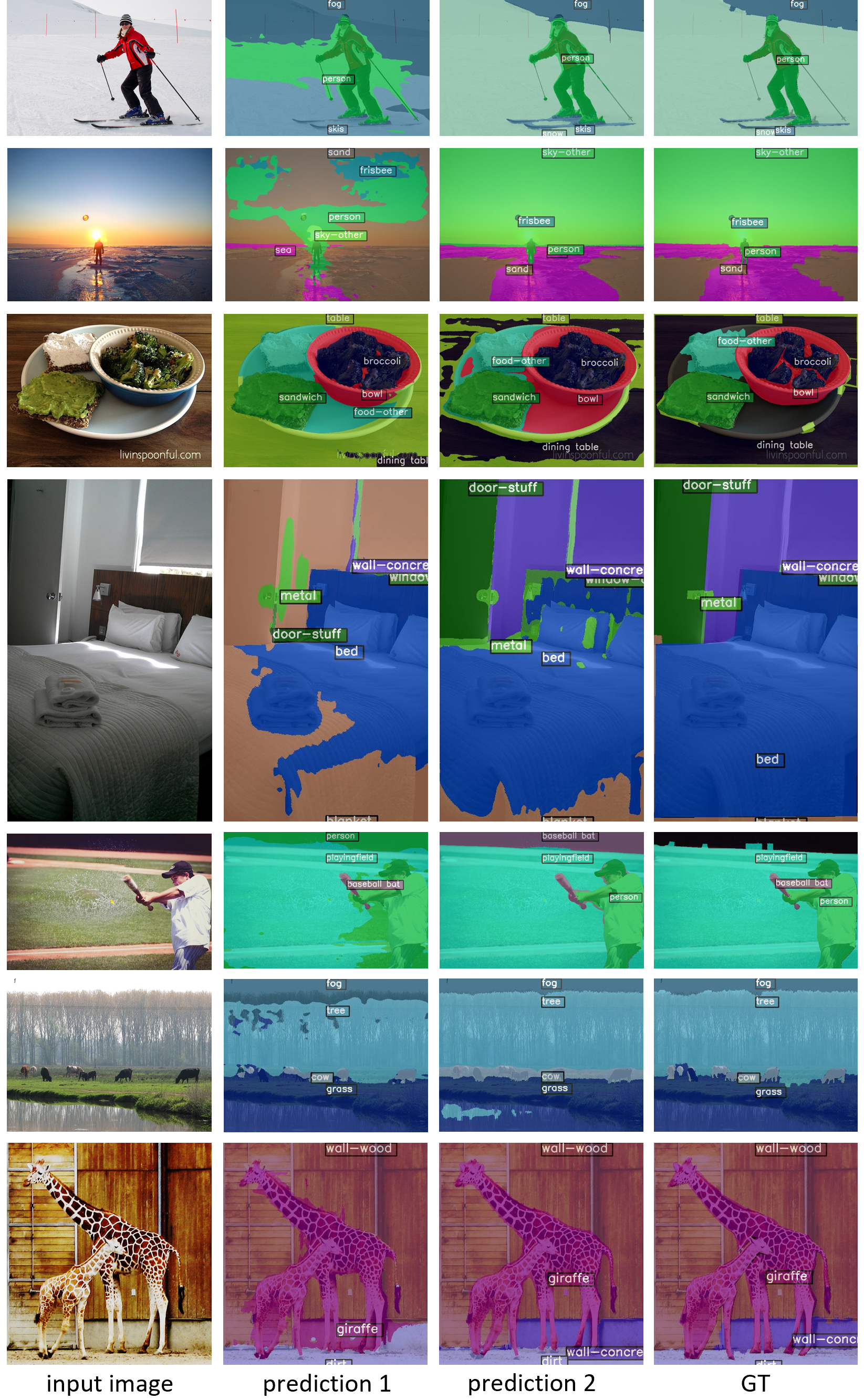}
    \caption{Visualization of segmentation maps on COCO-Stuff benchmark dataset.}
  \label{fig:sup-vis1}
\end{figure*}

\begin{figure*}[t]
  \centering
    \includegraphics[width=0.9\linewidth]{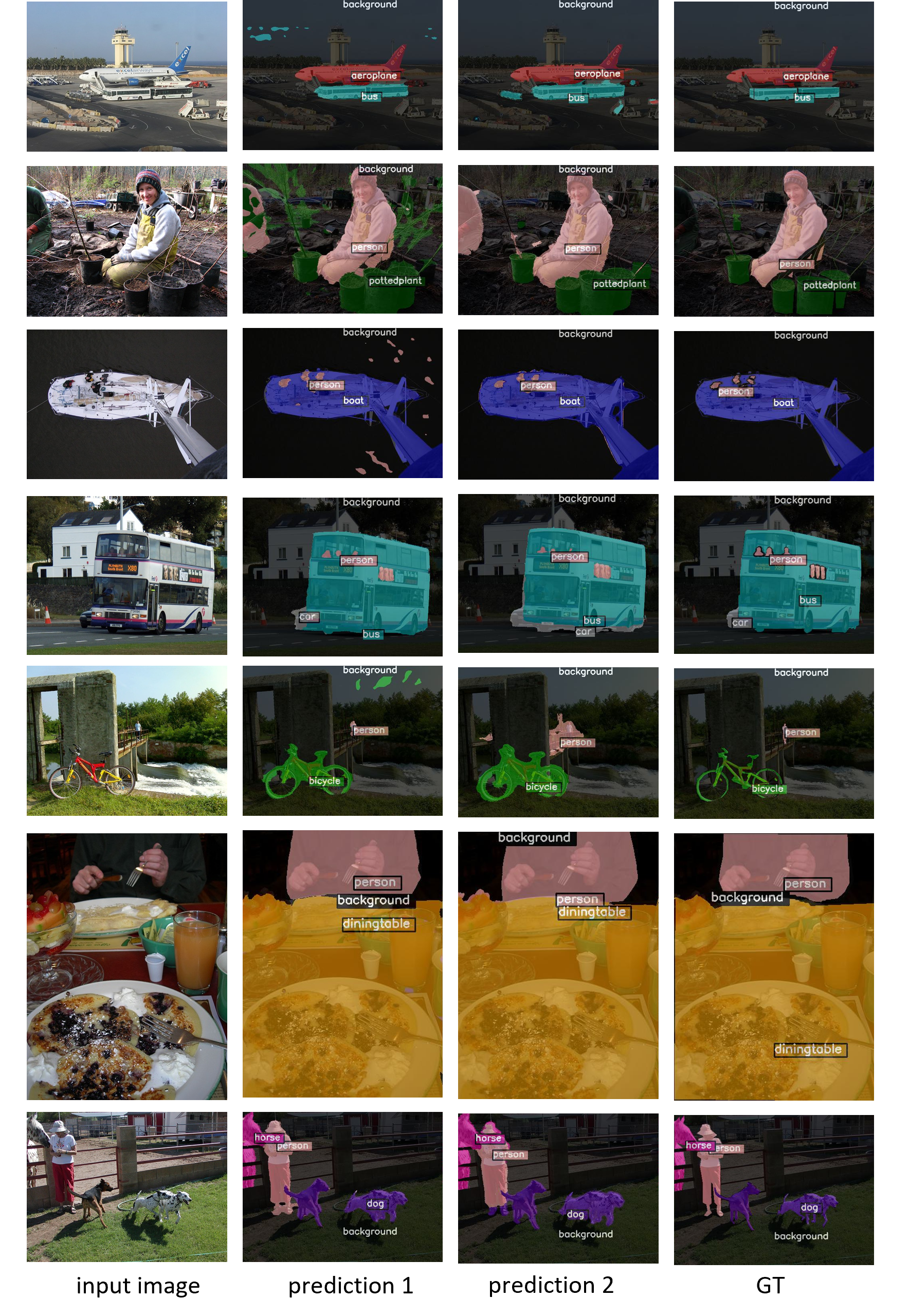}
    \caption{Visualization of segmentation maps on Pascal VOC benchmark dataset.}
  \label{fig:sup-vis2}
\end{figure*}

\begin{figure*}[t]
  \centering
    \includegraphics[width=0.9\linewidth]{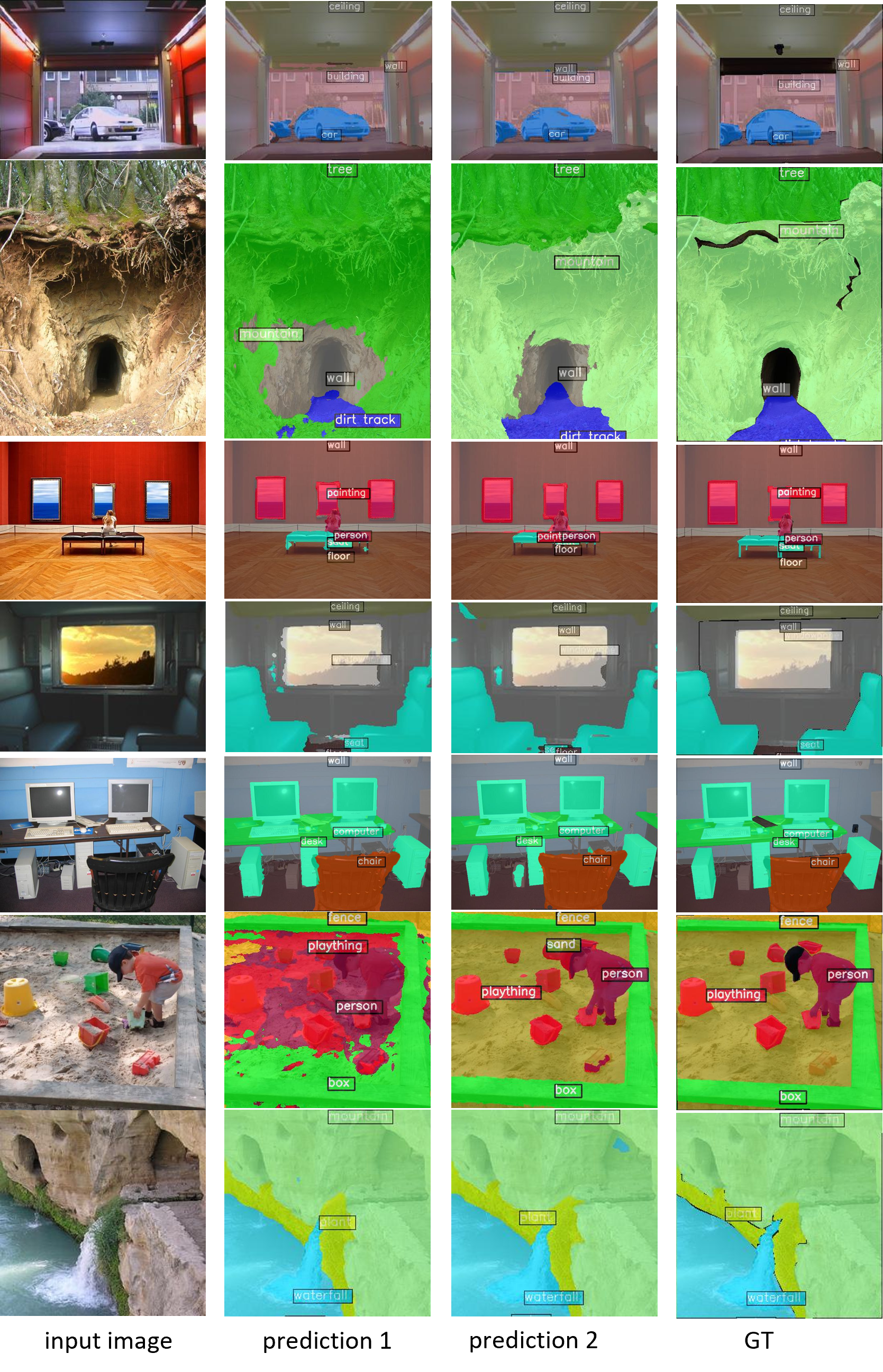}
    \caption{Visualization of segmentation maps on ADE150k benchmark dataset.}
  \label{fig:sup-vis3}
\end{figure*}

\begin{figure*}[t]
  \centering
    \includegraphics[width=0.9\linewidth]{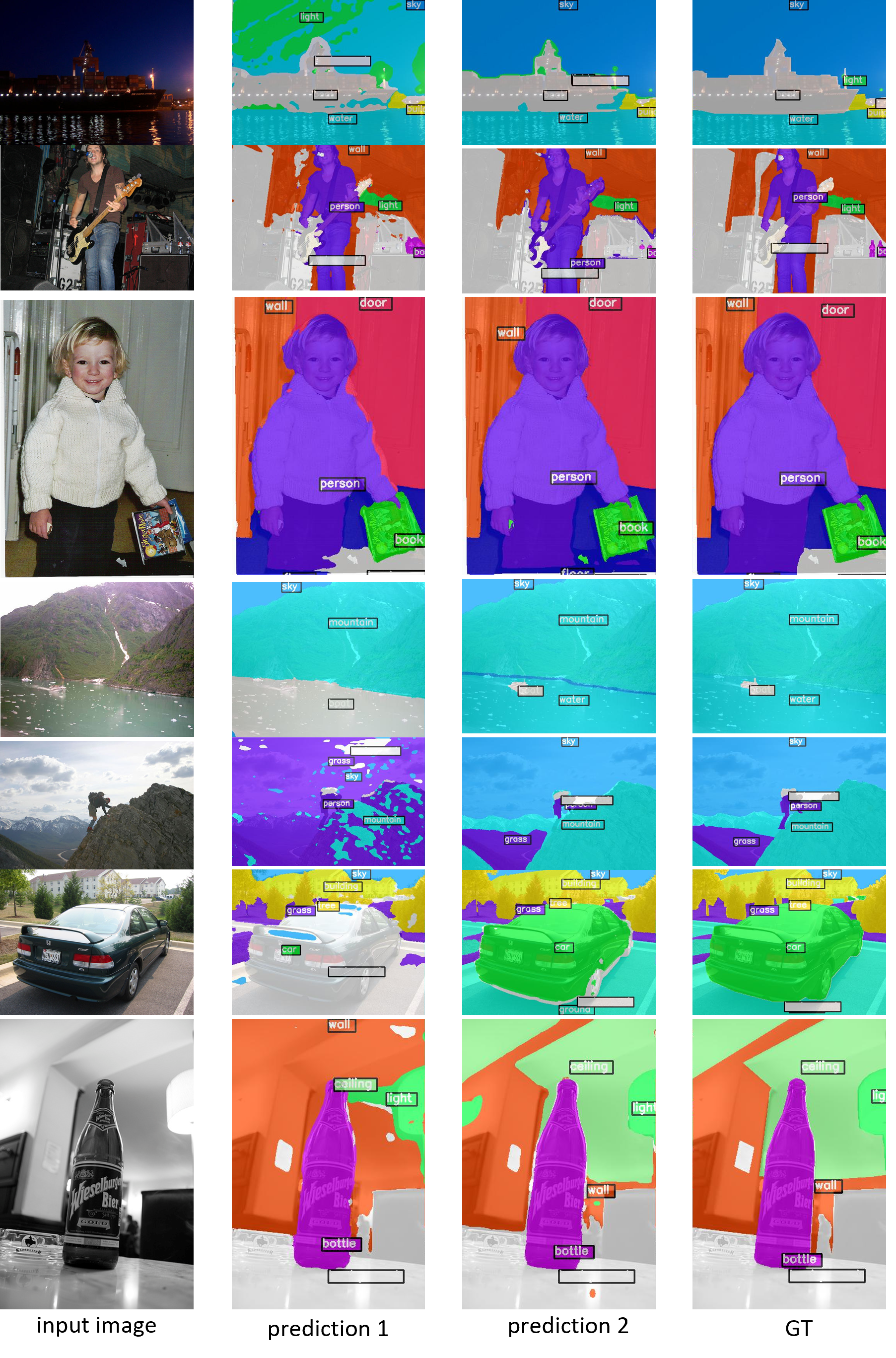}
    \caption{Visualization of segmentation maps on Pascal Context benchmark dataset.}
  \label{fig:sup-vis4}
\end{figure*}

\begin{figure*}[t]
  \centering
    \includegraphics[width=0.9\linewidth]{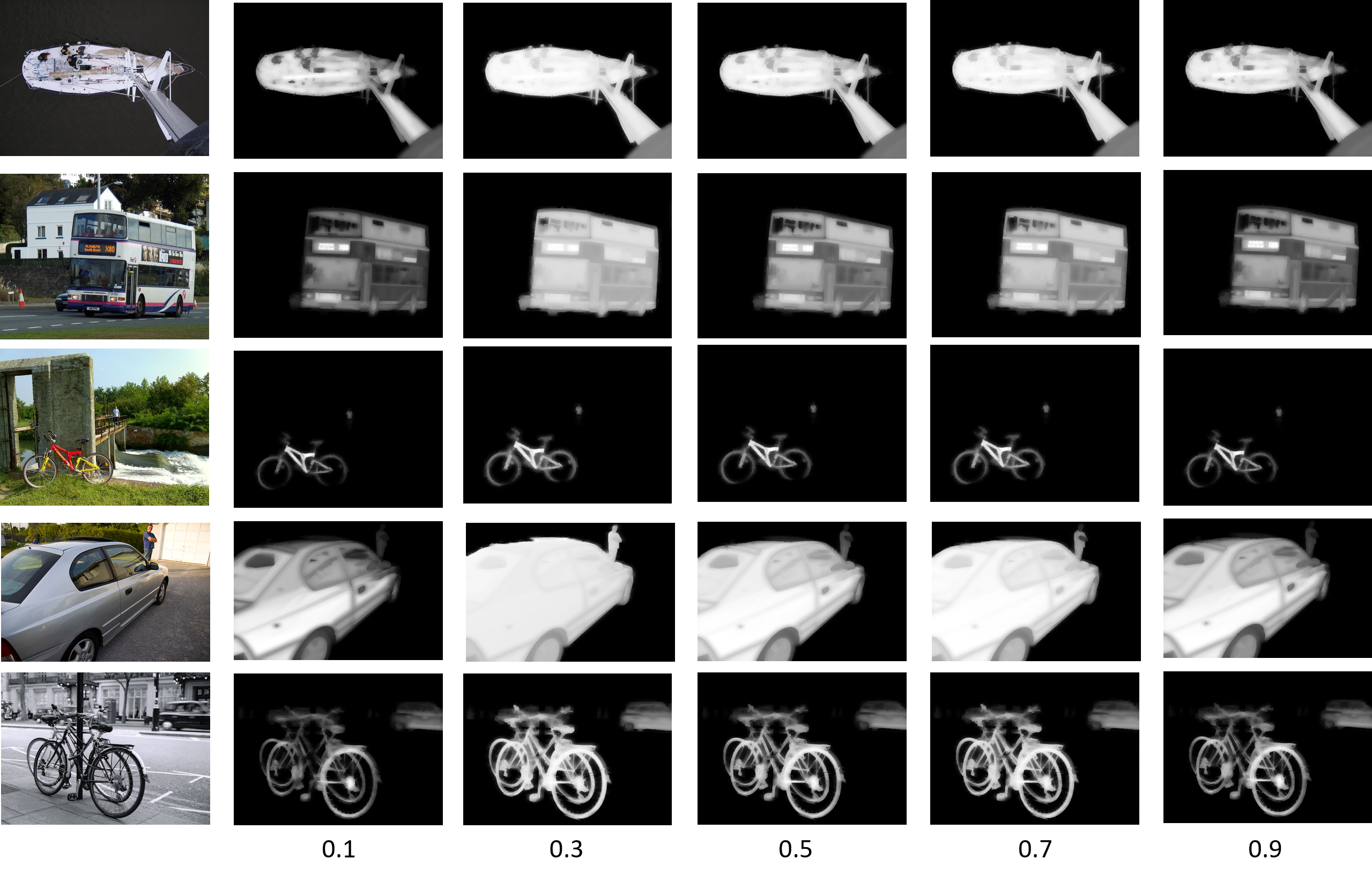}
    \caption{Visualization of maximum velocity maps across different threshold.}
  \label{fig:sup-vis5}
\end{figure*}

\begin{figure*}[t]
  \centering
    \includegraphics[width=0.9\linewidth]{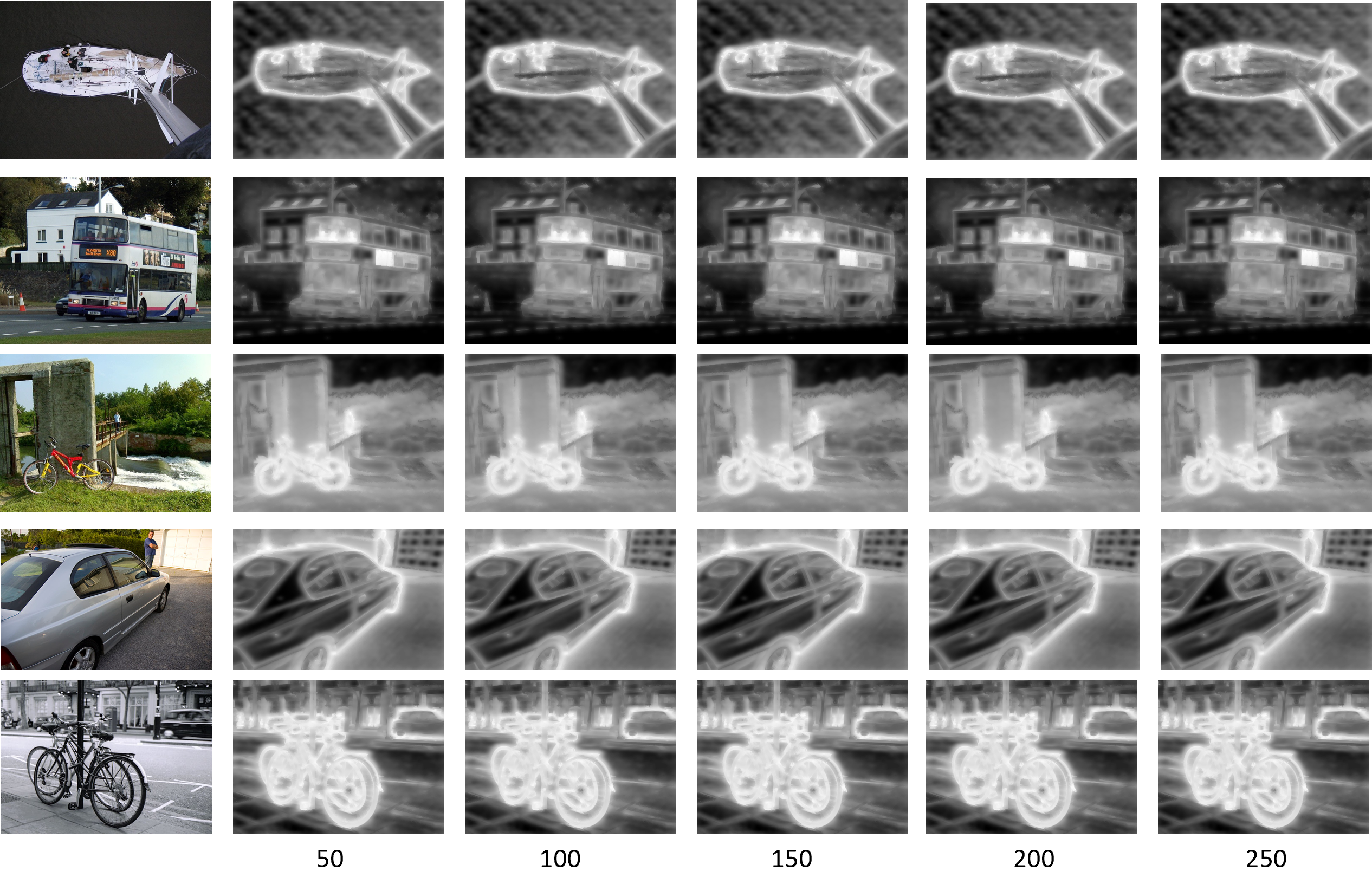}
    \caption{Visualization of optimal path maps across different iterations.}
  \label{fig:sup-vis6}
\end{figure*}

%% file: tables/sup_variant_comp.tex
\begingroup
\renewcommand{\arraystretch}{1.0}
\begin{table}[t]
\centering
\caption{Effect of different strategy for our components (unit: \%). \textbf{q-q mean}: the average attention tensor between queries across all layers. \textbf{k-k mean}: the average attention tensor between keys across all layers. All comparisons were performed using the optimal path mode.}
\label{tab:sup-variant_comp}
\resizebox{\linewidth}{!}{
    \begin{tabular}{ll!{\vrule height 10pt}ccccc}
    \toprule
    & \textbf{Variants} & \textbf{VOC21} & \textbf{COCO-Stuff} & \textbf{Cityscapes} & \textbf{ADE20K} & \textbf{Avg.}\\
    \midrule
    \midrule
    \multicolumn{7}{c}{self-attention weight combination ($\text{down}_0,\text{down}_1,\text{up}_0,\text{up}_1,\text{up}_2$)}\\
    \midrule
    \textbf{(I)} & ($1,0,0,0,0$) & 61.1 & 23.5 & 36.4 & 16.5 & 34.4 \\
    \textbf{(II)} & ($0,1,0,0,0$) & 62.2 & 25.1 & 38.2 & 17.6 & 35.8 \\
    \textbf{(III)} & ($0,0,1,0,0$) & 64.0 & 25.6 & 40.6& 18.3 & 37.1 \\
    \textbf{(IV)} & ($0,0,0,1,0$) & 65.1 & 26.4 & 41.5 & 20.8 & 38.5 \\
    \textbf{(V)} & ($0,0,0,0,1$) & 64.1 & 25.4 & 41.3& 19.2 & 37.5 \\
    \rowcolor{blue!5}
    \textbf{(VI)} & ($0,0,0.5,0.5,0$) & 66.9  & 28.6 & 41.7 & 22.8 & 40.0 \\
    \midrule
    \multicolumn{7}{c}{training-free logits-optimization methods}\\
    \midrule
    \textbf{(I)} & origin & 62.6 & 25.2 & 36.3 & 16.3 & 35.1 \\
    \textbf{(II)} & q-q mean & 64.3 & 26.2 & 41.2 & 20.5 & 38.1 \\
    \rowcolor{blue!5}
    \textbf{(III)} & k-k mean & 66.9  & 28.6 & 41.7 & 22.8 & 40.0 \\
    \bottomrule
    \end{tabular}}
\end{table}
\endgroup